%% file: main.tex
\PassOptionsToPackage{table,dvipsnames}{xcolor}
\documentclass[11pt]{fairmeta}
\setcitestyle{authoryear,round,citesep={;},aysep={,},yysep={;}}

\usepackage{microtype}
\usepackage{hyperref}
\usepackage{url}
\usepackage{booktabs}
\usepackage{textcomp}

\definecolor{darkblue}{rgb}{0, 0, 0.5}

\usepackage{latexsym}

\usepackage[T1]{fontenc}

\usepackage[utf8]{inputenc}
\DeclareUnicodeCharacter{2264}{\ensuremath{\leq}}

\usepackage{graphicx}
\usepackage{tabularx}
\usepackage{ragged2e}
\usepackage{multirow}
\usepackage{makecell}
\usepackage{xspace}
\usepackage{amsmath}
\usepackage{amsfonts}
\usepackage{array}
\usepackage[table,dvipsnames]{xcolor}
\usepackage{caption}
\usepackage{adjustbox}
\usepackage{enumitem}
\usepackage{xstring}
\usepackage{bbm}
\usepackage{pifont}
\usepackage{rotating}
\usepackage{wrapfig}
\usepackage{placeins}
\usepackage{fontawesome5}
\usepackage{XCharter}
\usepackage[xcharter,bigdelims,vvarbb]{newtxmath}
\usepackage[scaled=1.05,p]{zlmtt}

\DeclareTextFontCommand{\textbf}{\bfseries}
\newcolumntype{Y}[1]{>{\RaggedRight\arraybackslash}p{#1}}
\usepackage{siunitx}
\usepackage[most]{tcolorbox}
\usepackage{fvextra}
\DefineVerbatimEnvironment{VerbatimWrap}{Verbatim}{
  breaklines=true,
  breakindent=0pt,
  breaksymbol={},
  breakanywhere,
  fontsize=\small
}

\newcommand{\hl}[1]{\colorbox{ourGreen!15}{\strut #1}}

\usepackage{listings}
\usepackage{setspace}

\lstdefinestyle{caseboxstyle}{
  basicstyle=\ttfamily\small\setstretch{1}\selectfont,
  columns=fullflexible,
  keepspaces=true,
  showstringspaces=false,
  breaklines=true,
  moredelim=**[is][\bfseries\color{ourGreen}]{@@}{@@},
  escapeinside={(*@}{@*)},
  breakindent=0pt,
  breakautoindent=false,
}
\newtcolorbox{CaseBox}[1][]{
  enhanced,
  breakable,
  colback=ourSuperLightGreen,
  colframe=ourGreen,
  title=Error Case,
  fonttitle=\bfseries,
  subtitle style={font=\bfseries, colback=ourSuperLightGreen},
  left=6pt,right=6pt,top=6pt,bottom=6pt,
  segmentation style={solid, ourGreen, line width=0.8pt},
  #1
}

\definecolor{ourViolet}{HTML}{845EC2}
\definecolor{ourSuperLightViolet}{HTML}{F5F1F9}
\definecolor{ourLightViolet}{HTML}{B0A8B9}
\definecolor{ourDarkViolet}{HTML}{4B4453}
\definecolor{ourGreen}{HTML}{2A7F6F}
\definecolor{ourLightGreen}{HTML}{00C0A3}
\definecolor{ourSuperLightGreen}{HTML}{FAFFFD}
\definecolor{ourRed}{HTML}{B5523B}

\colorlet{metabg}{ourSuperLightViolet}
\colorlet{metafg}{ourDarkViolet}
\colorlet{metablue}{ourViolet}
\hypersetup{citecolor=ourGreen}

\newcommand{\ours}{\textsc{Mr Dre}\xspace}

\title{{\fontsize{18}{22}\selectfont Beyond Single-shot Writing: Deep Research Agents are Unreliable at Multi-turn Report Revision}}

\DeclareRobustCommand\afflogo[1]{\raisebox{0.6ex}{\includegraphics[height=1.6ex]{latex/figures/logos/#1}}}
\DeclareRobustCommand\affsep{\kern0.2em}
\DeclareRobustCommand\affmark[1]{\kern0.1em$^{#1}$}

\renewcommand\authorformat[2][]{\mbox{\fontsize{\authorsize}{14}\selectfont\sffamily\bfseries #2\kern0.15em#1}}
\newcommand\authorsize{11}

\author[\afflogo{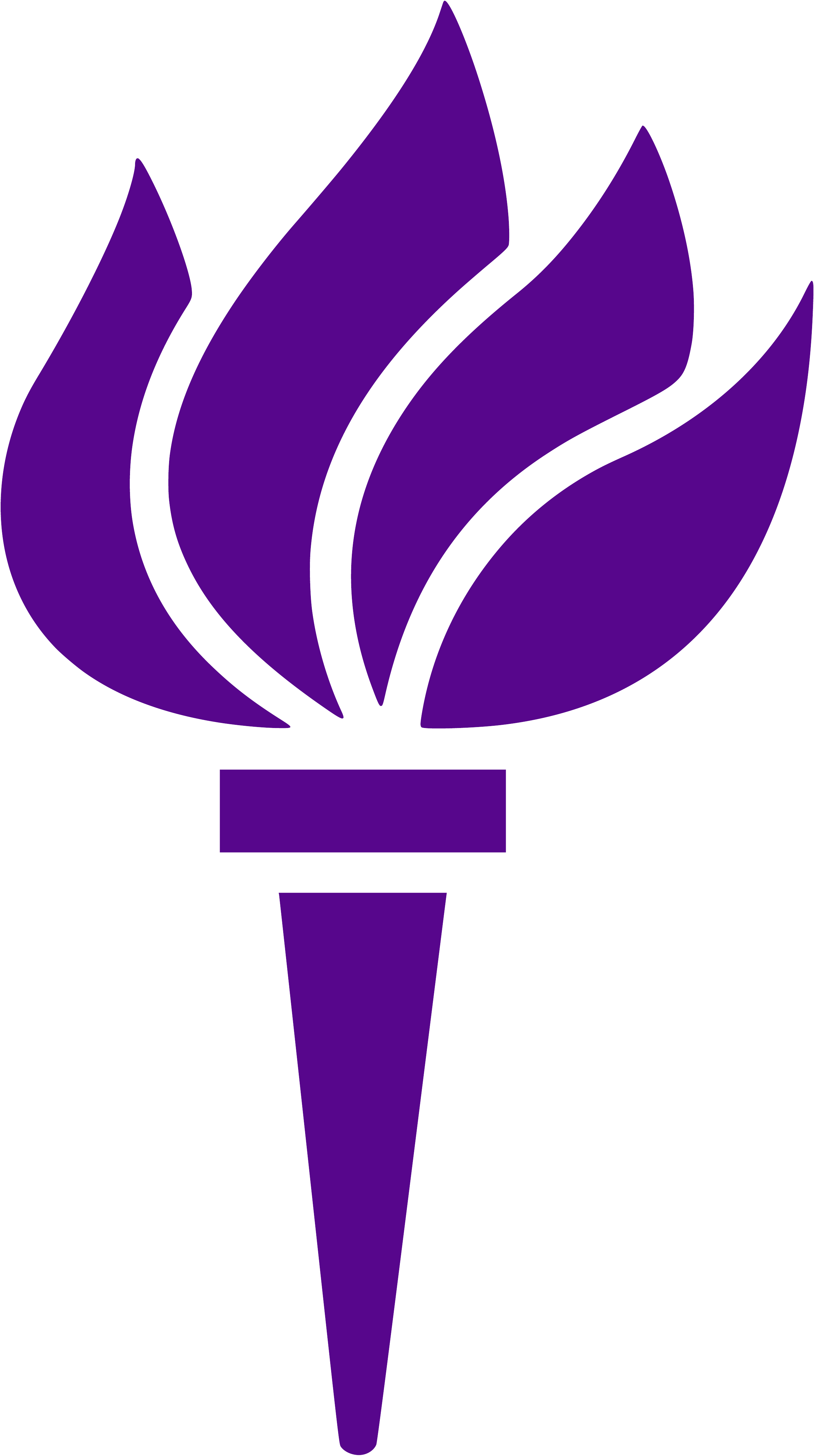}\affsep\afflogo{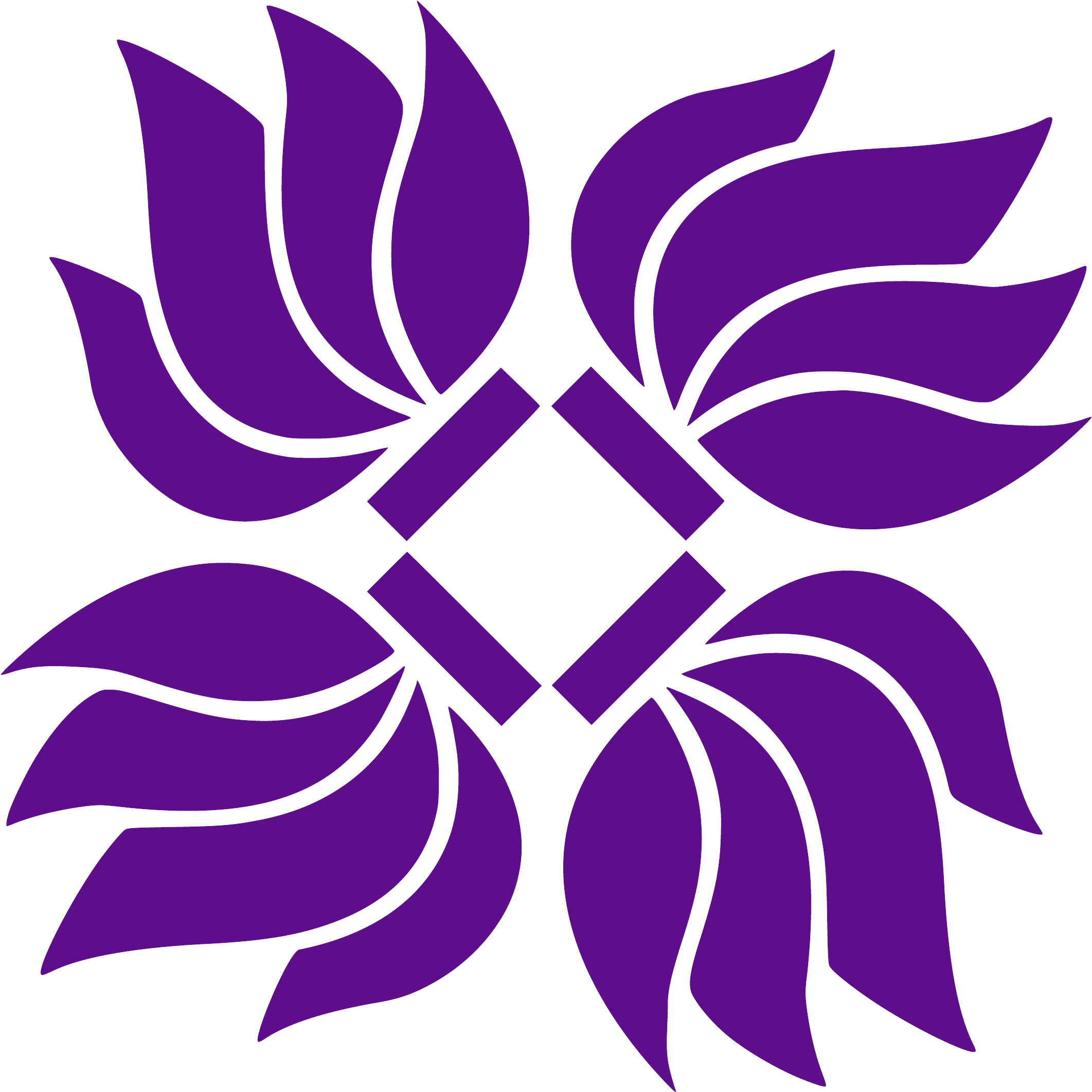}\affmark{*}]{Bingsen Chen}
\author[\afflogo{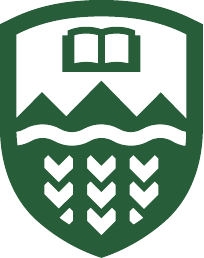}\affmark{*}\affmark{\dagger}]{Boyan Li}
\author[\afflogo{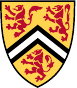}]{Ping Nie}
\author[\afflogo{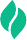}]{Yuyu Zhang}
\author[\afflogo{ualberta}\affsep\afflogo{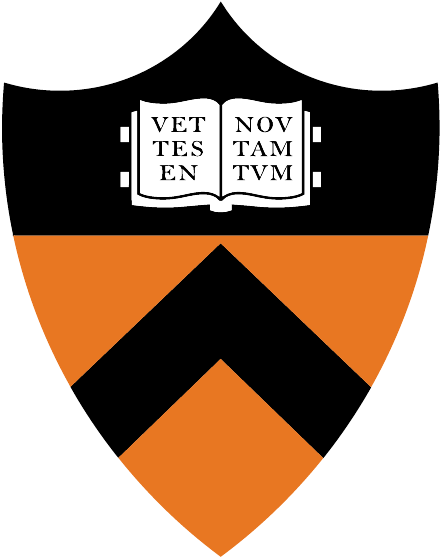}\affmark{\ddagger}]{Xi Ye}
\author[\afflogo{nyu}\affsep\afflogo{nyushanghai}\affmark{\ddagger}]{Chen Zhao}

\newcommand\affcell[2]{%
  \makebox[2.4ex][c]{$\vcenter{\hbox{\includegraphics[height=2.1ex]{latex/figures/logos/#1}}}$}\hspace{0.4em}#2}
\renewcommand\affiliationlist{%
  {\small
   \renewcommand{\arraystretch}{1.3}%
   \begin{tabular}{@{}l@{\hspace{1.6em}}l@{\hspace{1.6em}}l@{}}
     \affcell{nyu}{New York University} & \affcell{nyushanghai}{NYU Shanghai} & \affcell{ualberta}{University of Alberta} \\
     \affcell{princeton}{Princeton University} & \affcell{uwaterloo}{University of Waterloo} & \affcell{verdent}{Verdent AI, Inc.}
   \end{tabular}\par}%
  \vskip 0.15cm}

\date{January 19, 2026 (Last updated on \today)}
\metadata[Venue]{ACL 2026 Main Conference (\textcolor{ourViolet}{\smash{\scalebox{0.8}{\faTrophy}}~\textbf{SAC Highlight Award}})}
\metadata[Code and Data]{\url{https://github.com/BaleChen/Mr-Dre}}

\begin{document}

\abstract{

Existing benchmarks for Deep Research Agents (DRAs) treat report generation as a single-shot writing task, which fundamentally diverges from how human researchers iteratively draft and revise reports via self-reflection or peer feedback. Whether DRAs can reliably revise reports with user feedback remains unexplored. We introduce \ours, an evaluation suite that establishes multi-turn report revision as a new evaluation axis for DRAs.
\ours consists of (1) a unified long-form report evaluation protocol spanning comprehensiveness, factuality, and presentation, and (2) a human-verified feedback simulation pipeline for multi-turn revision.
Our analysis of five diverse DRAs reveals a critical limitation: while agents can address most user feedback, they also regress on 16--28\% of previously covered content and citation quality. Over multiple revision turns, even the best-performing agents leave significant headroom, as they continue to disrupt content outside the feedback's scope and fail to preserve earlier edits. We further show that these issues are not easily resolvable through inference-time fixes such as prompt engineering and a dedicated subagent for report revision.
}

\maketitle

{\renewcommand\thefootnote{}\footnotetext{$^{*}$Equal contribution.\quad $^{\dagger}$Work done while visiting NYU Shanghai.\quad $^{\ddagger}$Equal advising.}}

\input{latex/sections/1-Introduction}
\input{latex/sections/2-MR_DRE_v2}
\input{latex/sections/3-Experiments}
\input{latex/sections/4-Interventions}
\input{latex/sections/5-Related_Works}
\input{latex/sections/6-Conclusion}
\input{latex/sections/Limitations}
\input{latex/sections/Acknowledgements}

\bibliographystyle{plainnat}
\bibliography{custom}

\newpage
\beginappendix
\input{latex/sections/Appendix}

\end{document}

%% file: latex/sections/1-Introduction.tex
\input{latex/figures/teaser}

\section{Introduction}

Recent advances in the agentic capabilities of language models have led to the rise of Deep Research Agents (DRAs)~\citep{openai2025deepresearch, perplexity2025sonar, team2025tongyi, shao2025drtulu}. DRAs can tackle complex research queries by extensively searching and browsing the web, then synthesizing large volumes of information into long-form reports with rich citations and well-organized structure.

However, what dimensions to evaluate these complex systems against remains an open problem.
Early attempts~\citep{li2025webthinker, team2025tongyi} used multi-hop QA benchmarks~\citep{wei2025browsecomp, chen2025BrowseCompPlus, phan2025hle, mialon2023gaia} as proxies for evaluating DRAs' multi-step retrieval and reasoning ability. However, such benchmarks rely on short-form answers and fail to capture the report-writing capabilities of DRAs. More recent works have extended evaluation to long-form report generation~\citep{du2025deepresearch, yao2025rigorousbench, sharma2025researchrubrics, xu2025researcherbench}, but uniformly treat it as a \emph{single-shot} task: given a query, agents gather information and produce a report in one pass. This departs from human practice, where reports are produced through iterative revision, often guided by self-reflection or feedback.

In this work, we propose \textbf{multi-turn report revision} as a new evaluation axis for DRAs, where DRAs revise an initial report over multiple turns in response to user feedback. This capability is important for two reasons. First, in practical deployment, users often do not accept an initial response as-is~\citep{lee2025refinebench}, and may request additional details or changes in structure. Agents that cannot effectively incorporate such feedback while preserving the quality of the rest of the report provide limited utility. Second, iterative revision offers a natural mechanism for improving report quality with additional compute.
While test-time scaling has proven effective for reasoning~\citep{snell2025scaling, muennighoff2025s1}, its impact on report generation remains unexplored. We therefore ask: \textbf{Can current deep research agents reliably improve their reports via user feedback?}

To study this, we introduce \textbf{M}ulti-turn \textbf{R}evision of \textbf{D}eep \textbf{R}esearch Agent \textbf{E}valuation (\textbf{\ours}), a new evaluation suite supporting iterative report writing. For report evaluation, \ours unifies evaluation practices from prior benchmarks, which have adopted divergent metrics, into a lean protocol covering three dimensions: comprehensiveness, factuality, and presentation (Figure~\ref{fig:overview}, left). To enable multi-turn report revision evaluation, \ours provides a human-verified feedback simulation pipeline that generates realistic user feedback on report content and formatting (Figure~\ref{fig:overview}, right).

With \ours, we evaluate five diverse DRAs and find that current systems cannot reliably revise reports in response to user feedback. Although agents address over 90\% of requested edits, content or format revisions frequently reduce overall report quality: 16–28\% of previously covered content is broken, and citation quality degrades. Moreover, when extending revision to multiple turns, some agents show minimal to negative progress, and even the best-performing agents leave significant gaps as they continue to disrupt content outside feedback's scope and fail to preserve edits made in earlier turns. Finally, these failures persist despite inference-time remedies, including extensive prompt engineering and a dedicated reviser subagent. These findings suggest that reliable multi-turn revision will require more fundamental changes in DRA training and harness design, which we advocate as a priority for future research.

\noindent In summary, our main contributions are:
\begin{itemize}[noitemsep,topsep=0pt,leftmargin=*]

\item \textbf{A novel evaluation axis for DRAs.}
We identify multi-turn report revision as an important yet underexplored capability of DRAs.

\item \textbf{\ours evaluation suite.}
\ours unifies prior evaluation practices into a concise three-dimensional protocol for Deep Research report generation and includes a human-verified feedback simulation pipeline for multi-turn revision.

\item \textbf{Comprehensive analysis of current DRAs' multi-turn revision ability.} We reveal systematic limitations in current DRAs’ ability to revise reports from user feedback and show that inference-time fixes are insufficient.

\end{itemize}

%% file: latex/figures/teaser.tex
\begin{wrapfigure}{O}{0.5\textwidth}
  \centering
\includegraphics[width=0.5\textwidth]{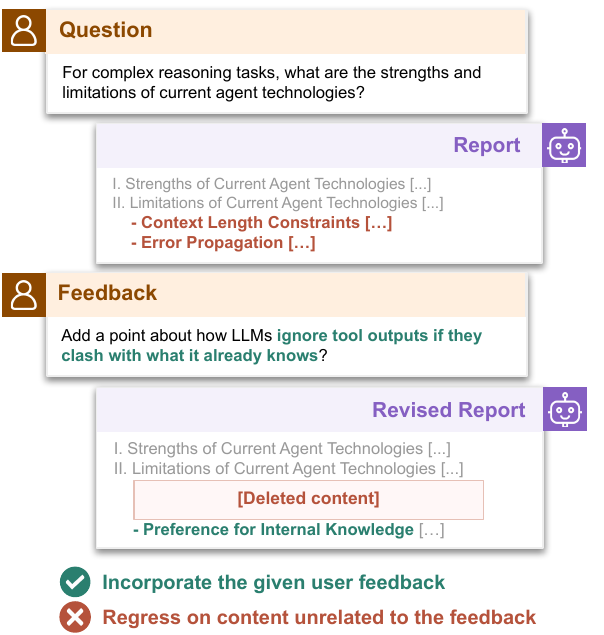}
  \caption{\textbf{Illustrative example of multi-turn revision failure in Deep Research Agents.} The revised report incorporates the user feedback but removes previously covered content that is outside the feedback's scope.}
\end{wrapfigure}

%% file: latex/sections/2-MR_DRE_v2.tex
\input{latex/figures/mrdre_flowchart}

\section{Task Definition} \label{sec:definitions}
We begin by formalizing deep research and defining our multi-turn report revision task:
Unlike standard search-augmented language models that are tasked with generating short-form~\citep{yang2018hotpotqa, wei2025browsecomp} or paragraph-long answers~\citep{fan2019eli5, han2024rag}, a Deep Research Agent is a system of one or multiple LLMs augmented with web searching tools that autonomously retrieves and analyzes vast online information and synthesizes findings into a comprehensive, well-cited research report.

Formally, given a user's initial query $q$, a DRA $\mathcal{A}$ generates a report $r_1 = \mathcal{A}(q)$. Current deep research benchmarks evaluate only this single output $r_1$, treating report writing as a one-shot task. We extend this paradigm to \textit{multi-turn report revision}. After receiving the initial report $r_1$, a user may provide feedback $f_1$, prompting the agent to revise the report and yield $r_2 = \mathcal{A}(q, r_1, f_1)$. This process can continue iteratively: at turn $t$, the agent produces $r_t = \mathcal{A}(q, r_1, f_1, \ldots, r_{t-1}, f_{t-1})$, conditioning on all previous turns of drafts and feedback.

\section{A Unified Deep Research Report Evaluation Protocol} \label{sec:protocol}
In this section, we introduce the first component of \ours, a comprehensive protocol for Deep Research report evaluation. We start with introducing our curated data (\S\ref{sec:data_curation}), and then detail the evaluation dimensions (\S\ref{sec:eval_dimensions}).

\subsection{Data Curation} \label{sec:data_curation}
To enable reliable evaluation, we guide our data curation with two criteria. First, each question must be paired with a \textit{question-specific rubric}: a set of content criteria (also referred to as rubric items) that a high-quality report should satisfy. Such rubrics provide ground-truth coverage targets tailored to each question and have proven effective for evaluating complex, long-form generations~\citep{ruan2025expertlongbench, lee2025checkeval}.
Second, the questions and rubrics must be sufficiently complex that DRAs are unlikely to satisfy all rubric criteria on their first attempt, leaving meaningful room for iterative revision.

Following these criteria, we include expert-annotated research questions from three datasets: ResearchRubrics~\citep{sharma2025researchrubrics}, Dr.~Bench~\citep{yao2025rigorousbench}, and ResearcherBench~\citep{xu2025researcherbench}. These datasets feature expert-annotated research-intensive questions paired with evaluation rubrics for improvement. Table~\ref{tab:data_stats} summarizes the dataset statistics.

\input{latex/tables/compare}

\subsection{Evaluation Dimensions \& Pipeline} \label{sec:eval_dimensions}

We combine best practices from prior benchmarks into three dimensions: \textit{comprehensiveness}, \textit{factuality}, and \textit{presentation}. These dimensions capture a minimal set of qualities that characterize an excellent research report: it should cover all essential information, make accurate citations and well-supported claims, and present content in a clear, well-organized manner.
We illustrate our evaluation protocol in Figure~\ref{fig:overview} (left), and compare \ours's unified protocol with previous benchmarks' evaluation in Table~\ref{tab:benchmark_comparison}. Note that our protocol is generally applicable to all Deep Research report generation tasks and can easily integrate new datasets.

\paragraph{Comprehensiveness} \label{sec:comprehensiveness}
A high-quality research report should cover all relevant aspects of a research question. We measure this using the coverage score of question-specific rubrics. A question $q$'s rubric $\mathcal{C}_q = \{(c_i, w_i)\}_{i=1}^{n}$ consists of $n$ criteria, where each criterion $c_i$ has a weight $w_i$ reflecting its importance. Following prior work~\citep{sharma2025researchrubrics, yao2025rigorousbench}, we adopt ternary grading to allow partial credit: an LLM judge $\mathcal{J}_\text{cov}$ evaluates each criterion against the question $q$ and report $r$, assigning a score $s_i = \mathcal{J}_\text{cov}(q, r, c_i) \in \{0, 0.5, 1\}$ corresponding to absent, partial, or complete coverage. The coverage score is the weighted average:
\[
    \textsc{Cov}(r) = \frac{\sum_{i=1}^{n} w_i \cdot s_i}{\sum_{i=1}^{n} w_i}.
\]
\paragraph{Factuality} \label{sec:factuality}
A high-quality research report should make accurate claims backed by reliable citations. Based on the in-line citations, we evaluate factuality from two angles: \emph{citation faithfulness} that measures the proportion of cited claims that are actually supported by their referenced sources, and \emph{claim groundedness} that measures the proportion of all claims that can be verified against external evidence. \looseness-1

Specifically, we adapt VeriScore~\citep{song2024veriscore} for Deep Research report evaluation. Given a report $r$, we first extract the set of atomic claims $\mathcal{E}(r) = \{e_1, e_2, \ldots, e_m\}$ using an LLM, where each claim $e_i$ is associated with zero, one, or multiple cited URLs. An LLM judge $\mathcal{J}_\text{fact}$ then evaluates each claim against the crawled content of its cited sources, classifying it as supported, contradicted, or insufficiently evidenced. We define citation faithfulness (left) and claim groundedness (right) as:
\[
    \textsc{Fa}(r) = \frac{|\mathcal{S}|}{|\mathcal{E}_{\text{cited}}(r)|}, \qquad
    \textsc{Gr}(r) = \frac{|\mathcal{S}|}{|\mathcal{E}(r)|}
\]
\noindent where $\mathcal{E}_{\text{cited}}(r) \subseteq \mathcal{E}(r)$ is the subset of claims with at least one citation, and $\mathcal{S} \subseteq \mathcal{E}(r)$ denotes the subset of supported claims. We provide additional technical details in Appendix~\ref{appendix:eval_fact}.

\paragraph{Presentation}
A high-quality research report should organize dense information into a readable, well-structured format with professional language. We consolidated and refined prior works'~\citep{fan2025understandingdeepresearchreports, yao2025rigorousbench, wang2025liveresearchbench} divergent criteria to arrive at a unified, static rubric of 10 questions shared across all research questions (listed in Table~\ref{tab:presentation_questions}). For each criterion $p_j$ in the rubric $\mathcal{P} = \{p_1, \ldots, p_{10}\}$, an LLM judge $\mathcal{J}_\text{pres}$ assigns a binary score. The overall presentation score is calculated as:
\[
    \textsc{Pre}(r) = \frac{1}{|\mathcal{P}|} \sum_{j=1}^{|\mathcal{P}|} \mathcal{J}_\text{pres}(r, p_j)
\]
\section{Extending Evaluation to Multi-turn Report Revision} \label{sec:extending_multiturn}

\ours also introduces an automated yet human-validated pipeline for multi-turn report revision evaluation. In this section, we introduce (1) a reliable way to simulate realistic user feedback (illustrated in Figure~\ref{fig:overview}, right), and (2) metrics to measure revision success.

\subsection{Feedback Simulation} \label{sec:feedback}

We design a feedback simulation pipeline that generates diverse and realistic user feedback on Deep Research reports. We consider three feedback categories corresponding to distinct revision settings: \underline{content}, \underline{format}, and \underline{self-reflection}.

\noindent\textbf{Content Feedback} requests adding new information or correcting existing content. Under this setting, a successful revision must address the feedback targets while preserving unrelated report content. To simulate such feedback, we leverage the rubric-based evaluation from \S\ref{sec:comprehensiveness}. Given a report draft, we first evaluate all rubric criteria using $\mathcal{J}_\text{cov}$, which we also ask for a justification for each score. We then uniformly sample $k$ uncovered criteria as \emph{feedback targets}, denoted $\mathcal{T}_t \subset \mathcal{C}_q$ for turn $t$, and prompt a feedback simulator LLM to generate natural feedback based on the question, the sampled feedback targets, and their corresponding scores and justifications. Grounding feedback in scoring justification rather than raw criteria text produces more natural requests that reflect how users would articulate what is missing.

\noindent\textbf{Format Feedback} targets the report’s structure, style, or presentation.
We expect DRAs to incorporate the formatting feedback without disrupting existing content coverage.
We curate 21 seed format feedback examples covering common user requests (e.g., adding bullet points, improving sectioning, or including TL;DR summaries. Full list in Table~\ref{tab:format_seeds}).
Given a report draft, we randomly sample three seed examples and prompt the LLM to select the most applicable one, then expand it into a piece of feedback specific to the report draft.
Using seed examples guides the simulator toward generating realistic feedback, while random sampling ensures diversity.

\noindent\textbf{Self-Reflection Feedback} provides no explicit revision guidance. The feedback is simply: \texttt{``Please reflect on your current report and revise it.''} This setting tests whether DRAs can autonomously identify and address deficiencies.

\noindent \textbf{Human Validation.}
We conduct human annotation to validate simulated content and format feedback along four dimensions: \textit{naturalness}, \textit{draft-specificity}, \textit{actionability}, and, for format feedback, \textit{content preservation}.
Our pipeline achieves near-ceiling scores across all dimensions with high inter-annotator agreement.  Details are in Appendix~\ref{appendix:feedback_sim}.

\subsection{Measuring the Success of Revision} \label{sec:metrics}

We introduce two additional metrics to measure the effectiveness of revision via the question-specific rubrics. Let $s_{i,t}$ denote the coverage score of criterion $c_i$ for report $r_t$.

\noindent \textbf{Incorporation Rate} measures whether the revision successfully incorporates feedback at turn $t$:
\[
    \textsc{Inc}= \begin{cases}
        \frac{1}{|\mathcal{T}_t|} \sum_{c_i \in \mathcal{T}_t} \mathbbm{1}\left[ s_{i,t} = 1 \right], & \text{Content feedback} \\[6pt]
        \mathcal{J}_\text{inc}(f_{t-1}, r_{t-1}, r_t), & \text{Format feedback}
    \end{cases}
\]
For content feedback, this is the proportion of feedback targets $\mathcal{T}_t$ that achieve full coverage after revision. For format feedback, an LLM judge $\mathcal{J}_\text{inc}$ assesses whether the formatting suggestion is followed (binary scoring).

\noindent \textbf{Break Rate} is the proportion of previously covered criteria whose scores degrade after revision:
\[
    \textsc{Brk} = \frac{|\{c_i : s_{i,t-1} > 0 \land s_{i,t} < s_{i,t-1}\}|}{|\{c_i : s_{i,t-1} > 0\}|}
\]
\noindent A break rate of 0 indicates that the revision preserves all previously covered content, while a high break rate suggests making destructive edits that fix one issue at the cost of disrupting others.

\noindent All metrics are defined per sample and reported as an average across the dataset. Prompt templates are in Appendix~\ref{appendix:prompt}.

%% file: latex/figures/mrdre_flowchart.tex
\begin{figure*}[t]
    \centering
    \includegraphics[width=\textwidth]{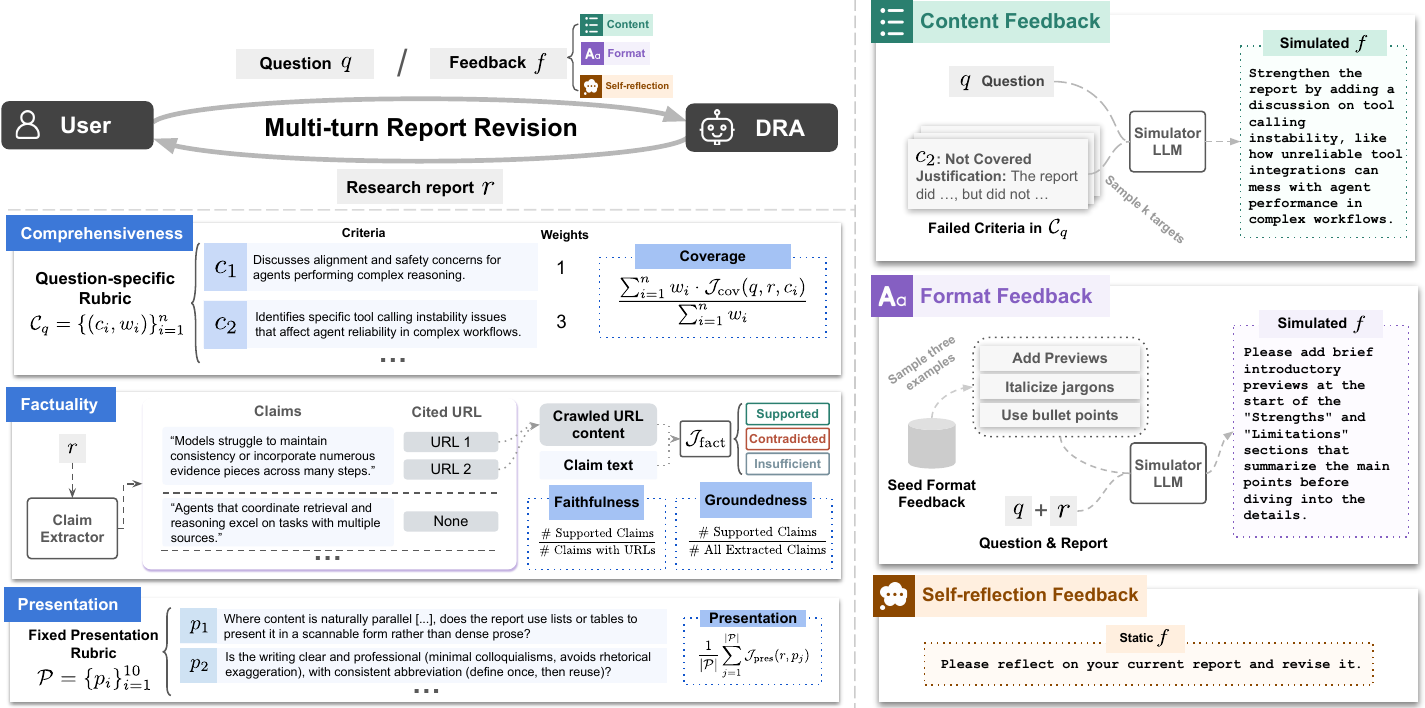}
    \caption{\textbf{\ours Evaluation Suite.} In multi-turn report revision, a DRA iteratively drafts and revises a report $r$ for question $q$ given user feedback $f$ (top left). \ours provides a unified Deep Research report evaluation protocol (bottom left) along three dimensions: Comprehensiveness, Factuality, and Presentation. To evaluate multi-turn revision performance, \ours provides a pipeline to simulate content, format, and self-reflection feedback (right).}
    \vspace{-0.5em}
    \label{fig:overview}
\end{figure*}

%% file: latex/tables/compare.tex
\newcommand{\cmark}{\ding{51}}%
\newcommand{\xmark}{\ding{55}}%
\newcommand{\greenCheck}{{\color{ourGreen}\cmark}}
\newcommand{\redCross}{{\color{ourRed}\xmark}}

\begin{table}[t]
\centering
\footnotesize
\setlength{\tabcolsep}{4pt}
\begin{tabular*}{\linewidth}{@{}l@{\extracolsep{\fill}}ccccc@{}}
\toprule
& \makecell{\textbf{Rubric-based}\\\textbf{Comprehensiveness}} & \makecell{\textbf{Claim-level}\\\textbf{Factuality}} & \makecell{\textbf{Report}\\\textbf{Presentation}} & \makecell{\textbf{Reference-free}\\\textbf{Evaluation}} & \makecell{\textbf{Multi-turn}\\\textbf{Revision}} \\
\midrule
ResearchRubrics~\citep{sharma2025researchrubrics} & \greenCheck & \redCross & \redCross & \greenCheck & \redCross \\
RACE\&FACT~\citep{du2025deepresearch} & \greenCheck & \greenCheck & \greenCheck & \redCross & \redCross \\
Dr.~Bench~\citep{yao2025rigorousbench} & \greenCheck & \redCross & \greenCheck & \greenCheck & \redCross\\
ResearcherBench~\citep{xu2025researcherbench} & \greenCheck & \redCross & \redCross & \greenCheck & \redCross\\
DR-ReportEval~\citep{fan2025understandingdeepresearchreports} & \redCross & \greenCheck & \greenCheck & \redCross & \redCross\\
DeepEval~\citep{wang2025liveresearchbench} & \greenCheck & \greenCheck & \greenCheck & \greenCheck & \redCross\\
\midrule
\ours & \greenCheck & \greenCheck & \greenCheck & \greenCheck & \greenCheck \\
\bottomrule
\end{tabular*}
\caption{\textbf{Comparison of \ours and previous Deep Research report evaluation protocols.} \ours checks all five aspects: rubric-based comprehensiveness evaluation, claim-level factuality assessment, report presentation scoring, requiring no reference answer, and support for multi-turn report revision (\S\ref{sec:extending_multiturn}).}
\label{tab:benchmark_comparison}
\end{table}

%% file: latex/sections/3-Experiments.tex
\input{latex/tables/main_table}

\section{Experiments}
Using \ours, we study our central research question: \textbf{how reliably can current DRAs revise reports via user feedback?} We first examine various DRAs' revision performance under three distinct feedback settings, and then analyze how revision behavior changes as we scale the number of turns (\S\ref{sec:multi-turn}) and feedback targets (\S\ref{sec:multi-item}).

\paragraph{Settings.} We evaluate five DRAs under three revision settings: self-reflection (Reflect), content feedback (Content$_k$), and format feedback (Format). In the Content$_k$ setting, each content feedback message targets $k$ rubric criteria that the previous-turn draft fails to satisfy. Due to the high cost of running proprietary DRAs, we conduct only the second-turn experiments on all three datasets in Table~\ref{tab:data_stats}. For experiments with up to 4 turns and varying the number of feedback targets ($k$), we use a sub-sampled Core Set (25 questions from each dataset, totaling 75 questions). Details about the subset creation are in Appendix~\ref{appendix:data_coreset}. Because Deep Research benchmarks are generally small, we report statistical significance tests for our main observations in Appendix~\ref{appendix:stat-sig}.

\paragraph{Evaluated Agents.}
We evaluate five DRA systems under three categories: \textbf{(1) Proprietary harness and models:} OpenAI o4-mini Deep Research (OpenAI DR)~\citep{openai2025deepresearch} and Sonar Deep Research by Perplexity (Sonar DR)~\citep{perplexity2025sonar}. Such systems reveal little information about the agents' details. \textbf{(2) Open harness, proprietary models:} LangChain Open Deep Research (LC ODR)~\citep{langchain2025odr}. It orchestrates a system of research, summarization, and finding compression agents, and a report-writing agent. We used GPT-4.1-nano for summarization and GPT-4.1-mini for the rest. \textbf{(3) Open harness and models:} Tongyi Deep Research (Tongyi DR)~\citep{team2025tongyi} and DR Tulu~\citep{shao2025drtulu}, which are post-trained for Deep Research report generation. Tongyi DR is not trained to generate citations, so we omit its citation-related results.

\subsection{Main Results}\label{sec:main_results}

Table~\ref{tab:eval-grid} shows the first and second turn results under three feedback settings. We observe that:

\noindent \textbf{DRAs struggle to reliably improve, or even preserve, report comprehensiveness across almost all feedback settings.} Across different DRAs, feedback types, and datasets, coverage scores predominantly decrease from Turn 1 to Turn 2. Under self-reflection, only LC ODR achieves a modest coverage gain (+3.6\%), while all other DRAs exhibit decreases or negligible changes. Even in the Content$_1$ setting, where the feedback explicitly identifies an unsatisfied rubric criterion, all agents except DR Tulu suffer coverage drops ranging from $-$2.0\% to $-$11.5\%. Format feedback, which by design should preserve content, leads to universal coverage drop across DRAs ($-$1.0\% to $-$16.2\%). Sonar DR, although achieving the best performance in the initial turn, also shows the largest performance drop across all feedback settings. These patterns indicate a systematic limitation in current DRAs' ability to revise reports based on different types of user feedback.

\noindent \textbf{While DRAs can follow most of the feedback instructions, they fail to preserve content outside the feedback's scope.} To understand why coverage degrades after revision, we examine the incorporation and break rates. All DRAs demonstrate strong instruction-following capabilities: incorporation rates mostly exceed 90\% for both content and format feedback. However, this success comes at the cost of disrupting previously satisfied content.
Break rates average 31\% under content feedback and 21\% under format feedback, as a substantial fraction of earlier coverage is lost after revision. Interestingly, break rates are lower under self-reflection, suggesting that more specific revision targets induce more aggressive edits that inadvertently affect unrelated content.

\noindent \textbf{Revision significantly degrades citation faithfulness and claim groundedness.} Beyond content coverage, factuality metrics also deteriorate after revision. The underlying causes vary across agents, such as fewer supported claims, and we provide a detailed analysis in Appendix~\ref{appendix:citation_full}.
Notably, Sonar DR exhibits the most severe degradation, especially after self-reflection, with faithfulness plummeting by up to $-$67.4\% and groundedness by up to $-$59.1\%. Further inspection reveals that Sonar DR produces reports with no citations 68\% of the time after receiving self-reflection feedback.

\input{latex/figures/multi_turn_all}

\subsection{How well can DRAs revise reports when extending to multiple turns?} \label{sec:multi-turn}

\noindent We extend revision up to four turns under the Content$_1$ and Reflect setting. We found that:

\noindent \textbf{DRAs fail to effectively accumulate coverage gains over multiple turns of content feedback.}
In the top row of Figure~\ref{fig:multi_turn_main}, we show each agent's coverage score (solid line) alongside the oracle score (dashed line), which represents the upper bound performance assuming perfect incorporation and zero break rate from Turn 1 onward. Tongyi DR and Sonar DR show minimal or even negative progress over multiple turns, while others achieve gradual coverage improvements with additional content feedback rounds. However, all agents lag far behind the oracle, and this gap shows no sign of closing over turns: by Turn 4, the gap between actual and oracle scores ranges from 9\% (OpenAI DR) to 26\% (Sonar DR).

As shown in the bottom two rows of Figure~\ref{fig:multi_turn_main}, the persistent gap from the oracle traces to both imperfect incorporation rates and non-trivial break rates across all agents. Notably, all DRAs break previously satisfied content at around 20-30\% by Turn 4. Tongyi DR and Sonar DR have consistently high break rates that offset any gains from incorporating content feedback over turns, whereas OpenAI DR, DR Tulu, and LC ODR show decreasing break rates (from 33\% to 21\% on average).

\noindent\textbf{DRAs break not only content outside of feedback's scope but also feedback targets from previous turns.} To measure this, we report the \textit{all-history incorporation rate}: the proportion of feedback targets from all previous turns that remain satisfied at turn $t$. While the current-turn incorporation rate stays stable around 90\%, all-history incorporation drops substantially. For instance, Sonar DR's all-history incorporation rate falls from 90\% at Turn 2 to 66\% by Turn 4. This gap indicates that agents fail to preserve previous fixes while addressing new feedback, even though earlier feedback remains in the input context.

\noindent We present additional multi-turn results in Appendix~\ref{appendix:mt-results}. Under the Content$_1$ setting, citation faithfulness and claim groundedness decrease over turns for all DRAs. Multiple turns of self-reflection similarly degrade coverage, citation faithfulness, and claim groundedness. These findings further underscore the unreliability of current DRAs in multi-turn report revision.

\noindent We present additional multi-turn results in Appendix~\ref{appendix:mt-results}. We found that citation faithfulness and claim groundedness decrease over turns across all DRAs under the Content$_1$ setting. Also, multiple turns of self-reflection similarly degrade coverage, citation faithfulness, and claim groundedness. These findings further underscore the unreliability of current DRAs in multi-turn report revision.

\subsection{How reliable are DRAs given feedback with multiple targets?} \label{sec:multi-item}

\input{latex/figures/multi_items_all}

We then examine how content feedback targeting multiple criteria affects revision performance.

\noindent\textbf{Increasing the number of feedback targets leads to higher coverage gains across all DRAs.} As shown in Figure~\ref{fig:multi_item}, all DRAs consistently achieve higher coverage as more unsatisfied criteria are given in the feedback. To understand this pattern, we examine the incorporation and break rates as $k$ increases: Incorporation rates remain high regardless of $k$, as agents can effectively handle multiple feedback targets at once. Interestingly, break rates also decrease with larger $k$, suggesting that agents tend to make less disruptive edits when given more targets to fix.

%% file: latex/tables/main_table.tex
\newcommand{\pha}{\phantom{00.00}}

\definecolor{turntwo}{gray}{0.96}

\newcommand{\sgn}[1]{\mbox{%
  \begingroup
  \IfSubStr{#1}{+}{\textcolor{ourGreen}{#1}}{%
    \IfSubStr{#1}{-}{\textcolor{ourRed}{#1}}{#1}%
  }%
  \endgroup
    }
}

\providecommand{\sysname}[1]{\parbox[c]{1.1cm}{\raggedright #1}}
\newcommand{\sysOpenAI}{\sysname{OpenAI\\DR}}
\newcommand{\sysPerp}{\sysname{Sonar\\DR}}
\newcommand{\sysOpen}{\sysname{LC\\ODR}}
\newcommand{\sysTongyi}{\sysname{Tongyi\\DR}}
\newcommand{\sysTulu}{DR Tulu}

\begin{table*}[t]
\centering
\fontsize{7.5}{7.5}\selectfont
\setlength{\tabcolsep}{1.9pt}
\renewcommand{\arraystretch}{1.15}

\begin{adjustbox}{max width=\textwidth}
\begin{tabularx}{\textwidth}{
    >{\raggedright\arraybackslash}p{1.1cm}
    >{\centering\arraybackslash}p{0.6cm}
    >{\centering\arraybackslash}p{1.3cm}
    *{4}{>{\centering\arraybackslash}X}
    @{\hspace{3pt}}!{\vrule width 0.6pt}@{\hspace{3pt}}
    *{4}{>{\centering\arraybackslash}X}
    @{\hspace{3pt}}!{\vrule width 0.6pt}@{\hspace{3pt}}
    *{4}{>{\centering\arraybackslash}X}
    @{\hspace{3pt}}!{\vrule width 0.6pt}@{\hspace{3pt}}
    *{2}{>{\centering\arraybackslash}X}
}
\toprule
\multicolumn{3}{l}{} &
\multicolumn{4}{c}{\textbf{ResearchRubrics}} &
\multicolumn{4}{c}{\textbf{Dr.~Bench}} &
\multicolumn{4}{c}{\textbf{ResearcherBench}} &
\textbf{Avg.} & \textbf{Avg.} \\
\cmidrule(lr){4-7}\cmidrule(lr){8-11}\cmidrule(lr){12-15}
\textbf{DRA} & \textbf{Turn} & \textbf{Type} &
\textbf{Cov.} & \textbf{Fa.} & \textbf{Gr.} & \textbf{Pre.} &
\textbf{Cov.} & \textbf{Fa.} & \textbf{Gr.} & \textbf{Pre.} &
\textbf{Cov.} & \textbf{Fa.} & \textbf{Gr.} & \textbf{Pre.} &
\textbf{Inc.} & \textbf{Brk.} \\
\midrule

& 1 & --
& 62.3 & 63.6 & 28.8 & 97.9
& 42.2 & 63.5 & 33.5 & 97.2
& 68.5 & 80.3 & 41.4 & 95.6
& -- & -- \\

\rowcolor{turntwo}\cellcolor{white} & \cellcolor{white} & Reflect
& \sgn{-1.5} & \sgn{-10.3} & \sgn{-8.4} & \sgn{+1.0}
& \sgn{-0.8} & \sgn{-2.4}  & \sgn{-4.4} & \sgn{+0.7}
& \sgn{-3.9} & \sgn{-7.0}  & \sgn{-5.3} & \sgn{+2.9}
& -- & 14.9 \\

\rowcolor{turntwo}\cellcolor{white} & \cellcolor{white} & Content$_1$
& \sgn{-6.6} & \sgn{-10.3} & \sgn{-6.1} & \sgn{-5.1}
& \sgn{-2.0} & \sgn{-16.9} & \sgn{-11.7} & \sgn{-5.4}
& \sgn{-9.1} & \sgn{-16.4} & \sgn{-11.0} & \sgn{-4.6}
& 93.6 & 29.7 \\

\rowcolor{turntwo}\cellcolor{white}\multirow{-4}{*}{\sysOpenAI} & \cellcolor{white}\multirow{-3}{*}{2} & Format
& \sgn{-2.2} & \sgn{-19.9} & \sgn{-13.0} & \sgn{+0.2}
& \sgn{-3.6} & \sgn{-8.8}  & \sgn{-8.4} & \sgn{0.0}
& \sgn{-3.9} & \sgn{-14.4} & \sgn{-7.1} & \sgn{+1.4}
& 98.5 & 14.8 \\

\midrule

& 1 & --
& 70.9 & 71.7 & 56.4 & 90.8
& 55.2 & 76.4 & 64.1 & 90.3
& 80.0 & 75.3 & 64.8 & 89.6
& -- & -- \\

\rowcolor{turntwo}\cellcolor{white} & \cellcolor{white} & Reflect
& \sgn{-5.2} & \sgn{-55.6} & \sgn{-48.1} & \sgn{+4.6}
& \sgn{-7.3} & \sgn{-60.8} & \sgn{-54.0} & \sgn{+2.5}
& \sgn{-5.2} & \sgn{-67.4} & \sgn{-59.1} & \sgn{+6.0}
& -- & 20.5 \\

\rowcolor{turntwo}\cellcolor{white} & \cellcolor{white} & Content$_1$
& \sgn{-11.5} & \sgn{-26.7} & \sgn{-19.7} & \sgn{+0.2}
& \sgn{-8.3}  & \sgn{-23.3} & \sgn{-21.9} & \sgn{-1.1}
& \sgn{-10.4} & \sgn{-8.5}  & \sgn{-3.9}  & \sgn{+0.6}
& 92.6 & 34.3 \\

\rowcolor{turntwo}\cellcolor{white}\multirow{-4}{*}{\sysPerp} & \cellcolor{white}\multirow{-3}{*}{2} & Format
& \sgn{-16.2} & \sgn{-35.6}  & \sgn{-28.8}  & \sgn{-6.1}
& \sgn{-13.9} & \sgn{-35.9}  & \sgn{-32.6}  & \sgn{-9.4}
& \sgn{-13.4} & \sgn{-30.0}  & \sgn{-27.1}  & \sgn{-3.1}
& 85.5 & 27.8 \\

\midrule

& 1 & --
& 60.9 & 72.5 & 39.7 & 99.5
& 43.0 & 74.2 & 46.1 & 99.8
& 71.0 & 82.6 & 51.8 & 99.8
& -- & -- \\

\rowcolor{turntwo}\cellcolor{white} & \cellcolor{white} & Reflect
& \sgn{+3.2} & \sgn{-4.8} & \sgn{-4.5} & \sgn{0.0}
& \sgn{+3.8} & \sgn{-3.3} & \sgn{-1.0} & \sgn{0.0}
& \sgn{+3.8} & \sgn{-8.5} & \sgn{-10.4} & \sgn{-0.2}
& -- & 8.5 \\

\rowcolor{turntwo}\cellcolor{white} & \cellcolor{white} & Content$_1$
& \sgn{-11.5} & \sgn{-6.5} & \sgn{-8.4} & \sgn{-1.3}
& \sgn{-5.3}  & \sgn{-5.2} & \sgn{-7.1} & \sgn{-1.9}
& \sgn{-8.7}  & \sgn{-9.1} & \sgn{-19.4} & \sgn{-0.5}
& 93.0 & 38.6 \\

\rowcolor{turntwo}\cellcolor{white}\multirow{-4}{*}{\sysOpen} & \cellcolor{white}\multirow{-3}{*}{2} & Format
& \sgn{-5.8} & \sgn{-1.4} & \sgn{-3.8} & \sgn{-1.9}
& \sgn{-3.5} & \sgn{-5.1} & \sgn{-8.6} & \sgn{-2.1}
& \sgn{-9.9} & \sgn{-4.2} & \sgn{-14.4} & \sgn{-2.4}
& 91.1 & 23.9 \\

\midrule

& 1 & --
& 58.5 &  --  &  --  & 99.3
& 39.4 &  --  &  --  & 99.2
& 68.4 &  --  &  --  & 99.9
& -- & -- \\

\rowcolor{turntwo}\cellcolor{white} & \cellcolor{white} & Reflect
& \sgn{-0.3} &  --  &  --  & \sgn{-1.2}
& \sgn{+0.2} &  --  &  --  & \sgn{+0.1}
& \sgn{-0.7} &  --  &  --  & \sgn{+0.2}
& -- & 9.9 \\

\rowcolor{turntwo}\cellcolor{white} & \cellcolor{white} & Content$_1$
& \sgn{-9.2} &  --  &  --  & \sgn{-5.3}
& \sgn{-0.9} &  --  &  --  & \sgn{-1.2}
& \sgn{-5.0} &  --  &  --  & \sgn{-0.8}
& 90.2 & 31.5 \\

\rowcolor{turntwo}\cellcolor{white}\multirow{-4}{*}{\sysTongyi} & \cellcolor{white}\multirow{-3}{*}{2} & Format
& \sgn{-6.7} &  --  &  --  & \sgn{-2.6}
& \sgn{-5.8} &  --  &  --  & \sgn{-2.5}
& \sgn{-9.8} &  --  &  --  & \sgn{-4.1}
& 94.3 & 23.2 \\

\midrule

& 1 & --
& 60.7 & 64.7 & 46.3 & 98.2
& 42.8 & 63.4 & 49.2 & 96.7
& 67.5 & 79.0 & 62.1 & 98.5
& -- & -- \\

\rowcolor{turntwo}\cellcolor{white} & \cellcolor{white} & Reflect
& \sgn{-0.7} & \sgn{-3.9} & \sgn{-0.4} & \sgn{-1.1}
& \sgn{-0.7} & \sgn{-2.6} & \sgn{+0.3} & \sgn{-1.9}
& \sgn{-0.1} & \sgn{-5.7} & \sgn{-3.5} & \sgn{-1.1}
& -- & 11.4 \\

\rowcolor{turntwo}\cellcolor{white} & \cellcolor{white} & Content$_1$
& \sgn{-2.7} & \sgn{-5.8} & \sgn{-2.6} & \sgn{-3.5}
& \sgn{+2.5} & \sgn{-4.0} & \sgn{-1.7} & \sgn{-1.1}
& \sgn{+1.0} & \sgn{-4.6} & \sgn{-6.8} & \sgn{0.0}
& 90.3 & 23.5 \\

\rowcolor{turntwo}\cellcolor{white}\multirow{-4}{*}{\sysTulu} & \cellcolor{white}\multirow{-3}{*}{2} & Format
& \sgn{-1.0} & \sgn{-2.2}       & \sgn{+2.2}       & \sgn{-1.3}
& \sgn{-2.2} & \sgn{+1.0}       & \sgn{+2.7}       & \sgn{-1.7}
& \sgn{-3.1} & \sgn{-5.7}       & \sgn{-2.6}        & \sgn{+0.7}
& 94.9 & 14.0 \\

\bottomrule
\end{tabularx}
\end{adjustbox}

\caption{\textbf{Main Results.} We report the Coverage (Cov.), Citation Faithfulness (Fa.), Claim Groundedness (Gr.), and Presentation (Pre.) score in percentage points. Incorporation (Inc.) and Break (Brk.) rate results are averaged across three datasets. For the second turn, we show the score changes from the first turn results for all evaluation metrics and feedback types, where improvement is colored in \textcolor{ourGreen}{green} and reduction is colored in \textcolor{ourRed}{red}.}
\label{tab:eval-grid}
\end{table*}

%% file: latex/figures/multi_turn_all.tex
\begin{figure}[t]
    \centering
    \includegraphics[width=\linewidth]{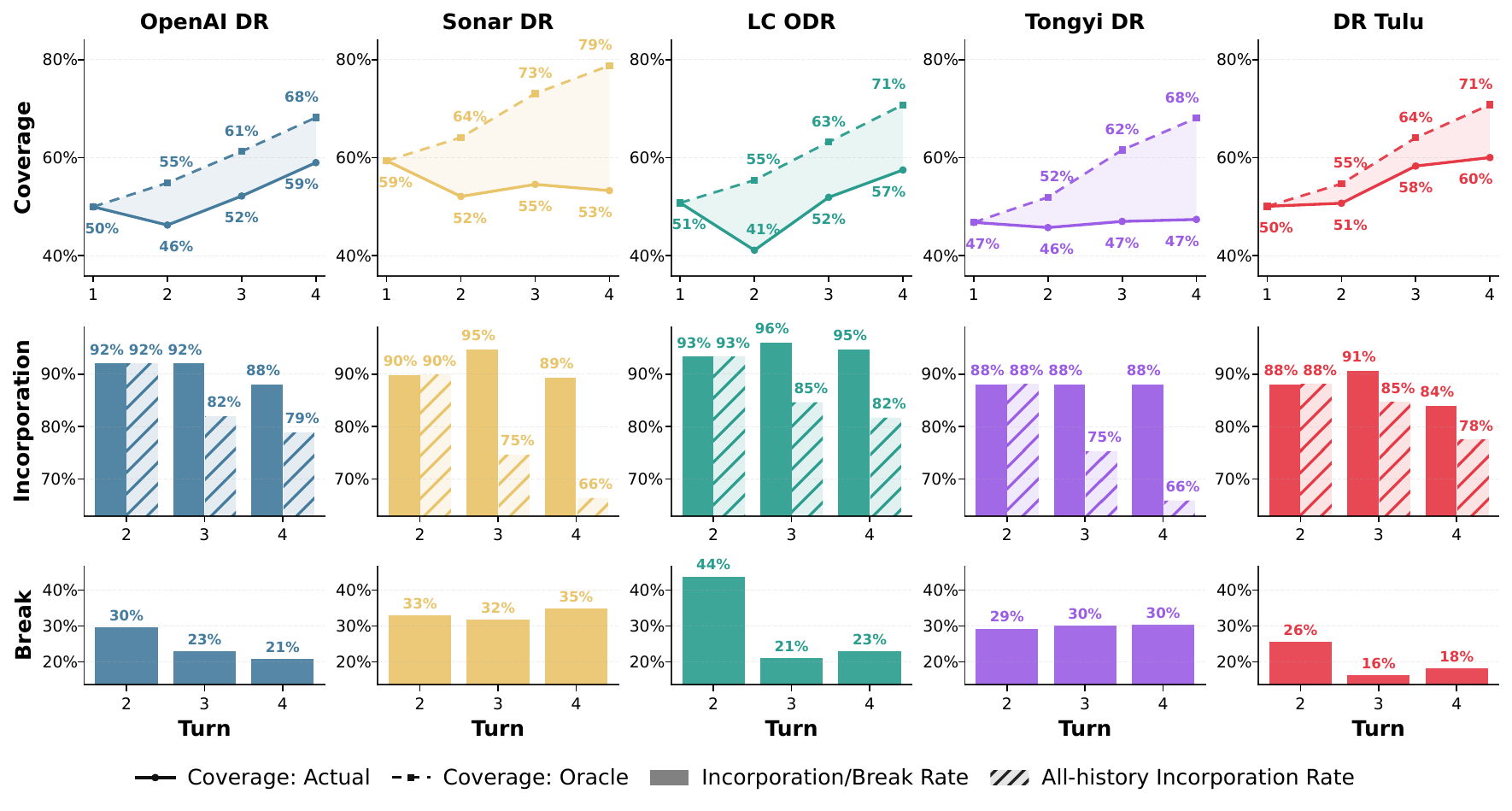}
    \caption{\textbf{Results for extending to 4 turns of revision under Content$_1$ setting}. We report the (top) coverage score (actual vs. oracle), (middle) incorporation rate, and (bottom) break rate.}
    \label{fig:multi_turn_main}
    \vspace{-1em}
\end{figure}

%% file: latex/figures/multi_items_all.tex
\newsavebox{\coresetbox}
\begin{figure}[t]
\centering
\sbox{\coresetbox}{\begin{minipage}{0.54\linewidth}\input{latex/tables/core_set_fix}\end{minipage}}%
\begin{minipage}[t]{0.43\linewidth}
    \centering
    \vspace{0pt}
    \includegraphics[height=\dimexpr\ht\coresetbox+\dp\coresetbox\relax,width=\linewidth,keepaspectratio]{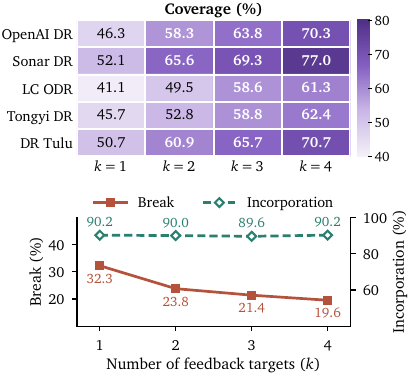}
    \captionof{figure}{\textbf{Effect of the number of feedback targets ($k$).} (Top) Coverage of each DRA. (Bottom) Break and incorporation rates averaged over the five DRAs, which share the same trend.}
    \label{fig:multi_item}
\end{minipage}%
\hfill
\begin{minipage}[t]{0.54\linewidth}
    \vspace{0pt}
    \usebox{\coresetbox}
    \captionof{table}{\textbf{PE and Reviser results.} Second turn's score change from Initial (turn 1) is shown for the four main metrics, along with incorporation (Inc.) and break (Brk.) rates. Best values are bolded within each agent and setting.}
    \label{tab:core-set-ablations}
\end{minipage}
\end{figure}

%% file: latex/tables/core_set_fix.tex
\centering
\fontsize{7.5}{7.5}\selectfont
\setlength{\tabcolsep}{2.8pt}
\renewcommand{\arraystretch}{1.08}

\begin{adjustbox}{max width=\linewidth}
\begin{tabularx}{\linewidth}{
    >{\raggedright\arraybackslash}p{1.1cm}
    >{\raggedright\arraybackslash}p{1.1cm}
    *{4}{>{\centering\arraybackslash}X}
    @{\hspace{3pt}}!{\vrule width 0.6pt}@{\hspace{3pt}}
    *{2}{>{\centering\arraybackslash}X}
}
\toprule
& &
\multicolumn{6}{c}{\textbf{Core Set}} \\
\cmidrule(lr){3-8}
\textbf{DRA} & \textbf{Setting} &
\textbf{Cov.} & \textbf{Fa.} & \textbf{Gr.} & \textbf{Pre.} & \textbf{Inc.} & \textbf{Brk.} \\
\midrule

\multirow{7}{*}{\sysOpenAI}
& Initial           & 50.0 & 74.6 & 37.7 & 97.0 & \multicolumn{1}{c}{--} & \multicolumn{1}{c}{--} \\
\cmidrule(lr){2-8}
& Content$_1$       & \sgn{-3.7} & \textbf{\sgn{-22.2}} & \sgn{-13.9} & \sgn{-3.5} & 91.9 & 29.6 \\
& +PE             & \sgn{+2.2} & \sgn{-26.3} & \sgn{-13.8} & \sgn{-1.8} & 93.2 & 16.1 \\
& +Reviser          & \textbf{\sgn{+5.1}} & \sgn{-30.4} &  \textbf{\sgn{-9.4}} & \textbf{\sgn{+1.0}} & \textbf{94.6} & \textbf{10.7} \\
\cmidrule(lr){2-8}
& Format            & \sgn{-4.5} & \sgn{-19.9} & \sgn{-12.6} & \textbf{\sgn{-0.3}} & 97.3 & 19.1 \\
& +PE             & \sgn{-3.7} & \sgn{-31.7} & \sgn{-17.3} & \textbf{\sgn{-0.3}} & \textbf{100.0} & \textbf{14.5} \\
& +Reviser          & \textbf{\sgn{-3.1}} & \textbf{\sgn{-16.3}} & \textbf{\sgn{-11.9}} & \sgn{-0.7} & 98.7 & 16.8 \\
\midrule

\multirow{7}{*}{\sysTulu}
& Initial           & 50.1 & 68.7 & 53.2 & 97.7 & \multicolumn{1}{c}{--} & \multicolumn{1}{c}{--} \\
\cmidrule(lr){2-8}
& Content$_1$       & \sgn{+0.6} & \sgn{-4.1} & \sgn{-4.0} & \sgn{-2.3} & 88.0 & 25.6 \\
& +PE             & \sgn{+3.4} & \textbf{\sgn{+1.3}} & \textbf{\sgn{+0.9}} & \sgn{-3.4} & 88.0 & 14.1 \\
& +Reviser          & \textbf{\sgn{+5.8}} & \sgn{-8.4} & \sgn{-11.3} & \textbf{\sgn{-2.1}} & \textbf{92.0} &  \textbf{9.5} \\
\cmidrule(lr){2-8}
& Format            & \sgn{-0.9} & \sgn{-1.8} & \textbf{\sgn{+1.5}} & \textbf{\sgn{-0.9}} & 94.0 & 13.1 \\
& +PE             & \sgn{-0.8} & \textbf{\sgn{-0.6}} & \sgn{+0.5} & \sgn{-1.5} & 96.0 & \textbf{12.1} \\
& +Reviser          & \textbf{\sgn{-0.3}} & \sgn{-1.4} & \sgn{-20.1} & \sgn{-1.3} & \textbf{100.0} & 12.6 \\

\bottomrule
\end{tabularx}
\end{adjustbox}

%% file: latex/sections/4-Interventions.tex
\section{Can Inference-time Fixes Improve Revision Performance?} \label{sec:intervention}

Our analysis reveals that DRAs cannot reliably revise reports based on user feedback, due to disruptive edits on existing content and citations, compounded by imperfect incorporation of the feedback. In this section, we investigate whether simple inference-time fixes can address these limitations without heavily modifying the underlying DRA system. Specifically, we test two approaches: \looseness-1

\noindent\textbf{Prompt Engineering (PE)} converts user feedback into a structured edit plan before revision. This approach decomposes the feedback into concrete, step-by-step instructions using an LLM (see example in Appendix~\ref{appendix:heavy-pe-example}), with the hypothesis that explicit guidance may help agents make more targeted edits without affecting unrelated content.

\noindent\textbf{Reviser Subagent (Reviser)} delegates the revision task to an LLM augmented with a web search tool. Since DRAs are optimized for multi-step tool calling and reasoning rather than localized editing, we hypothesize that a separate LLM with strong instruction-following capabilities can better incorporate user feedback while preserving content outside the feedback's scope.

\paragraph{Experimental Setup.} We evaluate the two fixes on OpenAI DR and DR Tulu under the Content$_1$ and Format settings using the Core Set. For PE, we use GPT-4.1 to transform simulated feedback into structured edit plans before sending them to the DRA. For the Reviser, we implement a ReAct agent~\citep{yao2023react} with Qwen3-30B-A3B-Instruct, a model with strong instruction-following capabilities (84.7\% on IFEval~\citep{zhou2023ifeval}), augmented with Google Search to retrieve additional information when needed. We include details and prompts in Appendix~\ref{appendix:intervention-details}.

\subsection{Findings}

\noindent\textbf{Both approaches can improve coverage by reducing break rates and improving incorporation rates.} As shown in Table~\ref{tab:core-set-ablations}, both PE and the Reviser enable DRAs to achieve coverage improvements for both content and format feedback settings, compared to when no fix is applied. These improvements stem from consistently higher incorporation rates and lower break rates. The Reviser generally outperforms PE in coverage scores, likely because the model is optimized for instruction-following and thus can execute edit requests more faithfully without introducing disruptive changes.

\noindent\textbf{However, these fixes fall short of fully addressing the challenges in multi-turn revision.} First, even with the best-performing Reviser, agents still break over 10\% of previously covered criteria on average, especially under the format feedback setting, where both fixes yield smaller gains. Second, neither approach resolves citation degradation: for OpenAI DR, both PE and the Reviser still show substantial drops in citation faithfulness and claim groundedness compared to the initial report. These persistent gaps indicate that inference-time mitigations alone cannot fully address the multi-turn revision challenges. Achieving reliable report revision that preserves both content coverage and citation quality will likely require more fundamental advances in training algorithms or harness design.

%% file: latex/sections/5-Related_Works.tex
\section{Related Work}

\paragraph{Deep Research Report Evaluation}
The emergence of DRAs has motivated long-form report benchmarking with varied evaluation approaches. Some rely on gold-standard reference reports to judge comprehensiveness~\citep{du2025deepresearch, li2025reportbench}, while others adopt rubric-based evaluation~\citep{hashemi2024llmrubrics, lee2025checkeval,arora2025healthbench} that specifies question-specific criteria to measure content coverage~\citep{wang2025liveresearchbench, yao2025rigorousbench, xu2025researcherbench, sharma2025researchrubrics}. A complementary axis concerns factual verifiability. Prior works assessed it via citation quality~\citep{gao2023alce, ye2024effective, liu2023evaluating}, which is widely adopted in recent Deep Research evaluations~\citep{fan2025understandingdeepresearchreports, yao2025rigorousbench, xu2025researcherbench, li2025reportbench}. Our \ours builds upon these evaluation practices to arrive at a unified protocol, meanwhile extending the scope to multi-turn report revision, an ability that remains underdeveloped for current DRAs.

\paragraph{Revision Abilities of LLMs}

Prior works have found that LLMs can improve their reasoning, coding, and agentic task performance through self-reflection~\citep{madaan2023self,zelikman2024self,shinn2023reflexion}. Yet, similar to our findings, some have also shown contradictory results that such gains can be fragile, as LLMs often fail to identify their own mistakes and thus struggle to self-correct reliably~\citep{huang2023large,lee2025refinebench}. Also, another line of work obtains feedback from external tools or critic models for LLMs to revise their outputs~\citep {gou2023critic,nathani2023maf,jiang2023active,wadhwa2024learningrefine}. In this work, we extend the discussion to Deep Research multi-turn report revision, examining both self-reflection and user feedback settings. Although \cite{qiao2025webresearcher} and \cite{han2025deep} explored iterative drafting for DRAs, they did not consider multi-turn user feedback settings, where we reveal critical limitations and provide a comprehensive testbed for future development.

%% file: latex/sections/6-Conclusion.tex
\section{Conclusion}
In this paper, we propose multi-turn report revision as an essential yet overlooked capability of DRAs. We introduce \ours, an evaluation suite featuring a unified evaluation protocol for long-form reports and a human-verified feedback generation pipeline for simulating user feedback in multi-turn revision. Across five diverse DRAs and three feedback settings, current systems cannot reliably improve reports through revision. While DRAs mostly address the given feedback, they frequently regress on unrelated content and citation quality. These gaps are not easily closed by simple fixes such as prompt engineering or dedicated subagents. We view multi-turn report revision as a critical missing piece in developing useful DRAs, and \ours aims to drive progress toward agents that can both conduct complex research and reliably adapt to users' evolving needs.

%% file: latex/sections/Limitations.tex
\section*{Limitations}

\paragraph{Understanding the Causes of Unreliability} While our work reveals critical limitations in DRAs' multi-turn revision ability, the causes of the high break rate, imperfect incorporation rate, and citation degradation are not yet fully understood (see error cases in Appendix~\ref{appendix:error-case}). We encourage future work to systematically analyze these underlying causes, which would inform the development of new training algorithms or agent harnesses for reliable multi-turn revision.

\paragraph{Model Scaling Effects on Revision Ability} Due to the high cost of running proprietary models on Deep Research tasks, we did not investigate how scaling up the backbone model affects revision reliability. For instance, we used o4-mini-deep-research instead of the stronger o3-deep-research for OpenAI DR, and LC ODR uses GPT-4.1-mini as its backbone. Although we provide a small-scale experiment with o3-deep-research in Appendix~\ref{appendix:scaling-o3}, how model scaling affects revision ability remains unclear and warrants more comprehensive investigation.

\paragraph{Missing Considerations in Evaluation Protocol} First, our feedback simulation assumes that the questions and rubrics are high-quality. Future work could enhance the feedback simulation pipeline to be robust to varying rubric quality, potentially incorporating LLM-based checklist evaluation~\citep{lee2025checkeval, wei2025rocketeval}. Second, real user feedback can be ambiguous or even contradictory, a complexity that \ours's feedback simulation does not yet capture. Since current DRAs already fail frequently when given concrete, actionable feedback, we expect such feedback to pose an even greater challenge, which makes it a promising direction for future work. Lastly, \ours does not penalize excessive report length. We observe that Sonar DR consistently achieves higher coverage, partially because its reports are substantially longer than those of other DRAs (on average 9452 tokens for Sonar DR vs 4516 tokens for other DRAs per report), though it also has a lower presentation score due to consistently failing $p_3$ in Table~\ref{tab:presentation_questions}. However, ideal length varies across questions and user preferences, making it difficult to define an evaluation scheme. Future work could build upon the \ours protocol to enhance length-aware evaluation.

%% file: latex/sections/Acknowledgements.tex
\section*{Acknowledgements}

We thank NYUSH NLP Lab for the helpful discussions. This work was supported in part through the NYU IT High Performance Computing resources, services, and staff expertise.

%% file: latex/sections/Appendix.tex
\section{Data Curation Details} \label{appendix:data}

\subsection{Dataset Details}
We show the dataset statistics of the three datasets of \ours in Table~\ref{tab:data_stats}.

\input{latex/tables/data_stats}

Note that for Dr.~Bench, we used their ``Query-Specific Rubrics'' as our question-specific rubrics, and excluded their coarser ``General-Report Rubrics'' in \cite{yao2025rigorousbench}. They labeled the score for satisfying each rubric criterion, which we use as the question weight. For all our experiments, we used a sub-sampled set of 100 questions in Dr.~Bench due to the high cost of Deep Research experiments, which is a sufficient size comparable to the other two datasets.

\subsection{Core Set Construction} \label{appendix:data_coreset}
We construct a core set by uniformly sampling 25 questions from each dataset (ResearcherBench, ResearchRubrics, and Dr.~Bench) such that for each sampled question, the initial report produced by every evaluated DRA fails to satisfy at least four evaluation criteria.

\section{Evaluation Protocol Details}
Below, we describe details of our evaluation protocol. For all LLM judges ($\mathcal{J}_\text{cov}$, $\mathcal{J}_\text{fact}$, $\mathcal{J}_\text{pres}$) and the claim extractor mentioned in \S\ref{sec:protocol}, we instantiate with GPT-4.1-mini (\texttt{gpt-4.1-mini-2025-04-14}) following DR Tulu, which is strong at instruction-following and long-context understanding, balancing accuracy and cost. We use temperature$=0$ for all judgements to minimize randomness and promote reproducibility.
\subsection{Comprehensiveness Evaluation} \label{appendix:eval_comp}
For comprehensiveness evaluation, we evaluate the coverage score using an LLM judge $\mathcal{J}_{\text{cov}}$ for each report-criterion pair. The prompt template is presented in Figure~\ref{fig:checklist_eval}.

Note that we append a small reminder text (Figure~\ref{fig:negative-reminder}) to the user message for evaluating negative-weight criteria. This is because the LLM Judge sometimes scores 1 when the report avoids what the criterion asks about, which is the opposite of what we expect it to do.

\begin{figure}[h!]
\begin{tcolorbox}[colback=ourSuperLightViolet,colframe=ourViolet,title=Negative-weight Reminder Text]
\begin{VerbatimWrap}
Note: this is a negative criterion. Your score should be 1.0 only if it describes something that is present or true to the report. If the report did not contain the content described by the criterion, your score should be 0.0.
\end{VerbatimWrap}
\end{tcolorbox}
\caption{\textbf{Negative-weight Reminder Text.}}
\label{fig:negative-reminder}
\end{figure}

\subsection{Factuality Evaluation} \label{appendix:eval_fact}

Here, we describe the detailed pipeline for factuality evaluation. We first split each report into sections using double new lines. Then, for each section, we extract atomic claims alongside their cited URL(s) with the claim extractor LLM (prompt template shown in Figure~\ref{fig:claim_extraction_sys_prompt}\&\ref{fig:claim_extraction_user_prompt}). For each claim, we gather its cited URLs and fetch the URL content using the Jina Reader API\footnote{\url{https://jina.ai/reader/}}. To reduce the cost of running LLM judges, we summarize the crawled URL content with a lightweight model, GPT-4.1-nano, to reduce context length. We then prompt $\mathcal{J}_\text{fact}$ to label each claim as \texttt{Supported}, \texttt{Contradicted}, or \texttt{Insufficient}, given the crawled URL content (Prompt in Figure~\ref{fig:supported_judge_sys_prompt}\&\ref{fig:supported_judge_user_prompt}). As introduced in \S\ref{sec:factuality}, we report citation faithfulness as the fraction of supported claims among claims with at least one URL, and claim groundedness as the fraction of supported claims among all extracted claims.

\subsection{Presentation Evaluation} \label{appendix:eval_presentation}
In Table~\ref{tab:presentation_questions}, we show the full list of our presentation questions, each with its detailed source or rationale for inclusion. These questions are carefully consolidated and refined from LiveResearchBench~\cite{wang2025liveresearchbench}, Dr.~Bench~\cite{yao2025rigorousbench}, DeepResearch-ReportEval~\cite{fan2025understandingdeepresearchreports}, with some new questions that we find missing from all prior works. Note that $p_6$ and $p_7$ are questions that might not apply to some reports; we exclude them if the judgement score is $-$1.

\input{latex/tables/presentation_questions}

\subsection{Handling Negative Weights in Evaluation Metrics}\label{appendix:eval_neg_weights}

Some Deep Research benchmarks, notably ResearchRubrics~\cite{sharma2025researchrubrics} in our evaluation suite, may assign negative weights to certain rubric criteria to penalize undesirable content such as misinformation or irrelevant topics. For a criterion $c_i$ with negative weight $w_i < 0$, a score of $s_i = 1$ (full coverage) indicates the report contains the undesirable content and should be penalized, while $s_i = 0$ indicates the report correctly avoids it. Below we describe how each metric accommodates negative weights. The actual results in our experiment sections use the formulations below instead of the simplified version in \S\ref{sec:protocol} without considering negative weights.

\subsubsection{Coverage Score}

For the coverage score, the numerator $\sum_{i=1}^{n} w_i \cdot s_i$ naturally handles negative weights: when $w_i < 0$ and $s_i > 0$, the product $w_i \cdot s_i$ is negative, reducing the overall score. However, the denominator must be adjusted to normalize only by the maximum achievable score, which comes from positive-weight criteria alone (since the best outcome for negative-weight criteria is $s_i = 0$, contributing nothing to the numerator). The full formula is:
\[
    \textsc{Cov}(r) = \frac{\sum_{i=1}^{n} w_i \cdot s_i}{\sum_{i: w_i > 0} w_i}
\]
This formulation ensures that the coverage score ranges from negative values (when the report contains penalized content) to 1 (when the report fully covers all positive-weight criteria and avoids all negative-weight criteria).

\subsubsection{Incorporation Rate}

For content feedback, the incorporation rate measures whether feedback targets reach their ideal coverage score after revision. The ideal score depends on the sign of the weight:
\[
    \bar{s}_i = \begin{cases}
        1, & \text{if } w_i > 0 \\
        0, & \text{if } w_i < 0
    \end{cases}
\]
For positive-weight criteria, the ideal is full coverage ($s_i = 1$). For negative-weight criteria, the ideal is zero coverage ($s_i = 0$), meaning the report should remove or avoid the undesirable content. The incorporation rate at turn $t$ becomes:
\[
    \textsc{Inc} = \frac{1}{|\mathcal{T}_t|} \sum_{c_i \in \mathcal{T}_t} \mathbbm{1}\left[ s_{i,t} = \bar{s}_i \right]
\]
Note that feedback targets $\mathcal{T}_t$ are sampled from criteria that have not yet reached their ideal score, which for negative-weight criteria means $s_{i,t-1} > 0$.

\subsubsection{Break Rate}

The break rate measures the degradation of previously achieved coverage. The definitions of "previously achieved" and "coverage degradation" are adapted based on the weight sign:

\paragraph{Previously Achieved Coverage} For positive-weight criteria, previously achieved coverage means $s_{i,t-1} > 0$ (at least partial coverage of desirable content). For negative-weight criteria, previously achieved coverage means $s_{i,t-1} < 1$ (not fully covering undesirable content, i.e., partially or fully avoiding the misconception). Let $\mathcal{C}^{+}_{t-1}$ denote the set of criteria with previously achieved coverage:
\[
\adjustbox{max width=\linewidth}{$
    \mathcal{C}^{+}_{t-1} = \{c_i : (w_i > 0 \land s_{i,t-1} > 0) \lor (w_i < 0 \land s_{i,t-1} < 1)\}
$}
\]

\paragraph{Coverage Degradation} For positive-weight criteria, degradation occurs when coverage decreases ($s_{i,t} < s_{i,t-1}$). For negative-weight criteria, degradation occurs when coverage increases ($s_{i,t} > s_{i,t-1}$), meaning the revision introduced more undesirable content. Both cases can be unified using the weight sign: degradation occurs when $w_i \cdot s_{i,t} < w_i \cdot s_{i,t-1}$, i.e., when the weighted contribution to the coverage score decreases. The full break rate formula is:
\[
    \textsc{Brk} = \frac{|\{c_i \in \mathcal{C}^{+}_{t-1} : w_i \cdot s_{i,t} < w_i \cdot s_{i,t-1}\}|}{|\mathcal{C}^{+}_{t-1}|}
\]

\subsection{Human Evaluation of Coverage LLM Judge} \label{appendix:human_cov_judge}

To further validate the reliability of the LLM judge and the break rate metric in \ours's evaluation protocol, we conduct a human evaluation. We randomly sample 30 rubric criteria that the judge scored as satisfied in Turn 1 but broken in Turn 2, drawn from all five DRAs' generations across the three datasets under both the Content$_1$ and Format feedback settings. This yields 60 (question, report, rubric criterion, LLM judgment) tuples in total. Two of our authors annotated whether they agreed with each judgment, scoring it as agree (1.0), partially agree (0.5), or disagree (0.0). Table~\ref{tab:llm_human_agreement_cov} reports the LLM--human agreement rate for Turn 1 and Turn 2 reports separately and combined.

\input{latex/tables/human_studies_cov_judge}

Agreement rates are high ($\geq$95\%) on both turns, confirming the reliability of our LLM judge and the break rate metric. Moreover, the similar agreement rates on Turn 1 and Turn 2 indicate that the judge does not behave differently on revised (potentially paraphrased) content than on the original.

\section{Feedback Simulation Pipeline Details}

\subsection{Content Feedback}
To simulate content feedback, we prompt GPT-4.1-mini with the question, $k$ sampled feedback targets with each score, weight, and scoring justification. We show the prompt for $k=1$ in Figure~\ref{fig:content1_sys_prompt} and $k>1$ in Figure~\ref{fig:contentk_sys_prompt}.

\subsection{Seed Format Feedback}
Two of our authors wrote the following 21 diverse and realistic seed format feedback pieces. We present them in Table~\ref{tab:format_seeds}.

\input{latex/tables/format_seeds}

\subsection{Human Validation Results}\label{appendix:feedback_sim}

Our goal is to simulate the most realistic follow-up that a human user would ask the DRA to revise the report against. Therefore, we defined the following four dimensions to assess the feedback's quality:

\vspace{.2em}
\noindent\textbf{Naturalness:} The language and wording should be natural and human-like, as if it were a natural follow-up response from the user themselves, or a thoughtful peer/supervisor.

\vspace{.2em}
\noindent\textbf{Draft-specificness:} The feedback should be tailored to the question and the current draft of the report, targeting aspects that the current draft misses and have clear room for improvement.

\vspace{.2em}
\noindent\textbf{Actionability:}The feedback should be concrete and actionable, phrased as implementable suggestions and avoiding vague comments such as “improve clarity” without explaining how.

\vspace{.2em}
\noindent\textbf{Content-preserving (only applicable to format feedback):} The feedback must not require any edits to existing content in the current draft. It should only incur changes in the form, structure, organization, tone, or style of the writing.

From all Content$_1$ and Format feedback generated for five DRAs across three datasets, we randomly sampled 50 content feedback instances and 50 format feedback instances. Two authors, each holding at least a Bachelor's degree in a science-related field, independently annotated each feedback instance alongside its corresponding report on the four dimensions above using binary scores (\texttt{satisfied} or \texttt{not satisfied}). We report agreement rate as the percentage of instances where both annotators assign identical scores across all four dimensions. For instances with disagreement, we take the lower score to provide a conservative estimate of feedback quality. We present the results in Table~\ref{tab:feedback_quality}.

\input{latex/tables/human_studies}

We found that our feedback simulation pipeline generally achieves a near-perfect score across all dimensions with a high inter-annotator agreement rate. This validates our feedback simulation pipeline as a realistic component for multi-turn report revision.

\section{Proposed Fixes Details} \label{appendix:intervention-details}

\subsection{Prompt Engineering (PE) on Feedback Details} \label{appendix:heavy-pe-example}
The prompt engineering (PE) fix pipeline refines raw user feedback into an executable revision instruction in two steps. First, we feed the original query, the full research report, and the user’s feedback into a prompt refiner (GPT-4.1) with a fixed system prompt (Figure \ref{fig:prompt_engi}) that forces the output into a structured, localized edit plan. Second, we append a fixed, hard-coded constraint suffix to this structured plan, which makes the downstream editor follow only the specified actions, avoid global rewrites, and output only the revised report. The concatenation of the structured edit plan and the constraint suffix (Figure \ref{fig:refiner_suffix}) forms the final refined prompt used for report revision.

\subsection{Reviser Subagent Details}
For the Reviser, we implemented a simple ReAct agent using Qwen3-30B-A3B-Instruct-2507 as the backbone model and its default function calling template, augmented with Serper API\footnote{https://serper.dev/} to call Google Search for additional information when the user feedback requires some extra information gathering. We set the temperature to 0.7, top-p to 0.95, and the maximum number of generated tokens to 16384. For each revision, we allow the agent to call the search API 10 times at maximum, with each call returning the top 5 web pages. If the maximal number of tool calls is reached, we softly force a final answer by using ``\texttt{You have reached the maximal number of web search calls. Please now produce the revised report based on the information you have and the user feedback.}'' as the tool output. The system and user prompt templates are in Figure~\ref{fig:reviser_sys} and \ref{fig:reviser_user}.

\section{Additional Results}

\subsection{Citation Analysis} \label{appendix:citation_full}
We present citation-related statistics in Table~\ref{tab:citation-stats}, reporting the average number of extracted claims ($|\mathcal{E}|$), claims with at least one citation ($|\mathcal{E}_\text{cited}|$), supported claims ($|\mathcal{S}|$), and citation counts ($|\mathcal{U}|$) for each dataset. These fine-grained statistics reveal that the causes of citation degradation vary across agents.

For \textbf{OpenAI DR}, degradation stems primarily from reductions in both supported claims and overall citation counts, with supported claims declining more severely (on average $-$7.0 supported claims and $-$11.2 citation counts).

For \textbf{Sonar DR}, as noted in Section~\ref{sec:main_results}, 68\% of reports generated after self-reflection contain zero citations, causing both factuality metrics to plummet. A similar pattern emerges in the Format setting, where 21\% of reports on average lack any cited URLs. While this phenomenon disappears for ResearchRubrics and ResearcherBench under Content$_1$, we still observe a substantial reduction in the ratio of supported claims.

For \textbf{LC ODR}, the Reflect setting produces notably more claims than the initial draft, yet the number of supported claims does not increase proportionally, leading to lower citation faithfulness and claim groundedness. In the Content$_1$ and Format settings, claim counts remain relatively stable or increase slightly, but the number of cited or supported claims drops, yielding similar degradation in citation quality.

For \textbf{DR Tulu}, the relatively stable but mixed number of supported claims across settings explains why it exhibits the most consistent citation quality among all evaluated DRAs.

\input{latex/tables/citation_full}

\subsection{Full Multi-turn Results under Content$_1$ and Reflect} \label{appendix:mt-results}
We present the complete results of Content$_1$ and Reflect up to 4 turns of revision in Figure~\ref{fig:multi_turn_content1} and Figure~\ref{fig:multi_turn_reflect}, respectively.

\input{latex/figures/multi_turn_4x5_appendix}

\subsection{Full Multi-item Content Feedback Results} \label{sec:full_multi-item}
We present the complete results of Content$_k$ with a varying number of feedback targets in Figure~\ref{fig:multi_item_full}.

\input{latex/figures/multi-tiem_5x6}

\subsection{Model Scaling Results with \texttt{o3-deep-research}} \label{appendix:scaling-o3}

Our main experiments use \texttt{o4-mini-deep-research} for OpenAI DR, although the presumably more capable \texttt{o3-deep-research} is available, and it is natural to ask how model scaling affects the multi-turn revision ability of DRAs. A full comparison is prohibitively expensive: a single \texttt{o3-deep-research} report costs an estimated \$5--8, and evaluating multi-turn revision requires three feedback settings and at least two turns of generation, which would cost up to \$2,400 even on our sub-sampled Core Set.

To nevertheless examine whether a stronger backbone model resolves the unreliability in multi-turn report revision, we run \texttt{o3-deep-research} on 10 randomly sampled questions from ResearchRubrics and report preliminary results in Table~\ref{tab:o3-results}.

\input{latex/tables/o3-results}

Even \texttt{o3-deep-research} exhibits coverage and citation degradation across the three feedback settings, with a notable 19.1\% break rate under content feedback. Because 10 samples yield noisy scores, these results are not directly comparable with Table~\ref{tab:eval-grid}. They nonetheless suggest that \emph{scaling the backbone model alone is unlikely to resolve the unreliability of multi-turn revision}.

\subsection{Statistical Significance Tests for Main Observations} \label{appendix:stat-sig}

Since Deep Research benchmarks are generally small, we conduct \emph{one-sided paired t-tests} on per-question differences to verify our main observations. For each question, the test takes the difference between two conditions (e.g., the coverage score before and after revision) and tests whether the mean difference is significantly below or above zero. We test three key observations and report the $p$-value for each DRA.\footnote{Throughout this section, * denotes $p<0.05$ and ** denotes $p<0.01$.}

\paragraph{Obs 1: Coverage decreases after revision in all feedback settings (Turn 1 $\rightarrow$ Turn 2).}
This trend appears on all three datasets in Table~\ref{tab:eval-grid}. We test $H_0$: coverage change $\geq 0$ against $H_1$: coverage change $< 0$ on results pooled across the three datasets (Table~\ref{tab:sig-obs1}).

\begin{table}[h!]
\centering
\fontsize{7.5}{7.5}\selectfont
\setlength{\tabcolsep}{2.8pt}
\renewcommand{\arraystretch}{1.08}
\begin{adjustbox}{max width=\linewidth}
\begin{tabularx}{0.6\linewidth}{
    >{\raggedright\arraybackslash}p{1.5cm}
    *{3}{>{\centering\arraybackslash}X}
}
\toprule
\textbf{Agent} & \textbf{Content$_1$} & \textbf{Format} & \textbf{Reflect} \\
\midrule
OpenAI DR   & $<$0.001** & $<$0.001** & $<$0.001** \\
Sonar DR    & $<$0.001** & $<$0.001** & $<$0.001** \\
LC ODR      & $<$0.001** & $<$0.001** & 1.0 \\
Tongyi DR   & $<$0.001** & $<$0.001** & 0.315 \\
DR Tulu     & 0.565 & 0.002** & 0.059 \\
\bottomrule
\end{tabularx}
\end{adjustbox}
\caption{$p$-values for Obs~1 (coverage decrease after revision).}
\label{tab:sig-obs1}
\end{table}

\paragraph{Obs 2: All DRAs exhibit significant break rates over multiple turns of revision.}
This trend appears in the bottom row of Figure~\ref{fig:multi_turn_main}. We test $H_0$: break rate $\leq 0$ against $H_1$: break rate $> 0$ under the Content$_1$ setting (Table~\ref{tab:sig-obs2}).

\begin{table}[h!]
\centering
\fontsize{7.5}{7.5}\selectfont
\setlength{\tabcolsep}{2.8pt}
\renewcommand{\arraystretch}{1.08}
\begin{adjustbox}{max width=\linewidth}
\begin{tabularx}{0.6\linewidth}{
    >{\raggedright\arraybackslash}p{1.5cm}
    *{3}{>{\centering\arraybackslash}X}
}
\toprule
\textbf{Agent} & \textbf{Turn 2} & \textbf{Turn 3} & \textbf{Turn 4} \\
\midrule
OpenAI DR   & $<$0.001** & $<$0.001** & $<$0.001** \\
Sonar DR    & $<$0.001** & $<$0.001** & $<$0.001** \\
LC ODR      & $<$0.001** & $<$0.001** & $<$0.001** \\
Tongyi DR   & $<$0.001** & $<$0.001** & $<$0.001** \\
DR Tulu     & $<$0.001** & $<$0.001** & $<$0.001** \\
\bottomrule
\end{tabularx}
\end{adjustbox}
\caption{$p$-values for Obs~2 (break rate over multiple turns).}
\label{tab:sig-obs2}
\end{table}

\paragraph{Obs 3: Actual coverage lags behind the oracle over multiple turns of revision.}
This trend appears in the top row of Figure~\ref{fig:multi_turn_main}. We test $H_0$: actual $-$ oracle $\geq 0$ against $H_1$: actual $-$ oracle $< 0$ under the Content$_1$ setting (Table~\ref{tab:sig-obs3}).

\begin{table}[h!]
\centering
\fontsize{7.5}{7.5}\selectfont
\setlength{\tabcolsep}{2.8pt}
\renewcommand{\arraystretch}{1.08}
\begin{adjustbox}{max width=\linewidth}
\begin{tabularx}{0.6\linewidth}{
    >{\raggedright\arraybackslash}p{1.5cm}
    *{3}{>{\centering\arraybackslash}X}
}
\toprule
\textbf{Agent} & \textbf{Turn 2} & \textbf{Turn 3} & \textbf{Turn 4} \\
\midrule
OpenAI DR   & $<$0.001** & $<$0.001** & $<$0.001** \\
Sonar DR    & $<$0.001** & $<$0.001** & $<$0.001** \\
LC ODR      & $<$0.001** & $<$0.001** & $<$0.001** \\
Tongyi DR   & 0.003** & $<$0.001** & $<$0.001** \\
DR Tulu     & 0.022* & $<$0.001** & $<$0.001** \\
\bottomrule
\end{tabularx}
\end{adjustbox}
\caption{$p$-values for Obs~3 (actual coverage vs.\ oracle).}
\label{tab:sig-obs3}
\end{table}

\medskip

\noindent All three observations are statistically significant at $p < 0.01$ for almost all agents, confirming that our findings reflect systematic limitations of current DRAs in multi-turn revision.

\section{Prompt Templates} \label{appendix:prompt}
We present prompt templates in Figure \ref{fig:content1_sys_prompt}-\ref{fig:reviser_user}.

\section{Case Studies}
\subsection{More Feedback Examples}
We present representative feedback examples used in our multi-turn revision setup (Table \ref{tab:more_feedback_examples}). The table includes both format feedback and content feedback with one or three targets.
\input{latex/tables/feedback_example}

\subsection{Error Cases} \label{appendix:error-case}
We provide two representative failure cases in multi-turn report revision. Figure~\ref{fig:example_prompt_missing_content} shows a \emph{missing content} case, where the revised report fails to preserve content outside the feedback's scope and drops a required paragraph. Figure~\ref{fig:example_missing_facts} illustrates \emph{citation degradation}, where the revised report reduces citations and removes in-context citation markers from the original.

\onecolumn
\input{latex/figures/examples/case_example}

\input{latex/figures/examples/content_feedback_prompt}
\input{latex/figures/examples/evaluation_prompt}
\input{latex/figures/examples/refine_system_prompt}
\input{latex/figures/examples/refiner_suffix}
\input{latex/figures/examples/reviser_prompt}

%% file: latex/tables/data_stats.tex
\begin{table}[h]
\centering
\footnotesize
\setlength{\tabcolsep}{5pt}
\renewcommand{\arraystretch}{1.1}

\begin{tabular}{
    l
    c
    c
    l
}
\toprule
\textbf{Dataset} & \textbf{Size} & \makecell{\textbf{\# Criteria}\textbf{/ Question}} & \textbf{Domains} \\
\midrule
ResearchRubrics   & 101 & 25.67 & General \\
Dr.~Bench    & 214 & 14.32 & General \\
ResearcherBench  &  65 & 13.33 & AI\&ML \\
\bottomrule
\end{tabular}

\caption{\textbf{Dataset statistics}. \# Criteria / Question denotes the average number of question-specific rubric criteria per question.}
\label{tab:data_stats}
\end{table}

%% file: latex/tables/presentation_questions.tex
\begin{table}[t]
\centering
\small
\renewcommand{\arraystretch}{1.18}
\begin{tabular}{@{}c p{8cm} p{5cm}@{}}
\toprule
 & \textbf{Question} & \textbf{Source/Rationale} \\
\midrule
$p_1$ & Does the report follow a clear, logically ordered structure that is easy to navigate (e.g., problem $\rightarrow$ approach $\rightarrow$ results), with sections that match the report's stated purpose and directly address the research question? & Q1 in LiveResearchBench's Table 3; GRR 1 and 2 in Dr.~Bench; Definition of ``Clear and Logical Structure'' in DeepResearch-ReportEval\\
$p_2$ & Do different sections logically follow or build on one another with minimal redundant restatement, and is any repetition clearly purposeful (e.g., brief recap before a new stage)? & Definition of ``Redundancy'' in DeepResearch-ReportEval; Q2 in LiveResearchBench's Table 3; Refined so that recap/summary is not counted as redundancy\\
$p_3$ & Where content is naturally parallel (steps, criteria, comparisons, key takeaways), does the report use lists and/or tables to present it in a scannable form rather than dense prose? & Newly written. This is not present in any previous evaluations, but is essential for penalizing dense paragraphs without proper formatting. \\
$p_4$ & Are headings/subheadings consistent in level and hierarchy (H1/H2/H3), and are comparable sections named with parallel phrasing (e.g., ``Method,'' ``Results'' rather than inconsistent mixes like ``How they did it,'' ``Findings'')? & Further specified GRR8, 42 in Dr.~Bench. Added that the heading names should be parallel and comparable. \\
$p_5$ & Does the report use concise transition sentences/phrases to signal why the subsequent content follows and to reduce abrupt jumps and make the report easier to follow? & Further specified  GRR7 in Dr.~Bench and Definition of ``Clear and Logical Structure'' in DeepResearch-ReportEval.\\
$p_6$ & If there are cross-references, are they consistent and unambiguous (figure/table numbers, section references, in-text citation), with no missing/duplicate numbering and no ``see above/below'' without an anchor? If no cross-references are present, the score should be $-$1. & Extended Q3, 4, 6, 10 in LiveResearchBench's Table 3 and GRR24 in Dr.~Bench to all types of cross-references. \\
$p_7$ & If tables are included, are they structurally complete and interpretable on their own (no blank cells without notation, consistent units/precision, clear headers/labels/notes)? If no tables are included, the score should be $-$1. & Extended Q8 in LiveResearchBench's Table 3; We do not allow automatic pass but rather discard it if the report does not have tables\\
$p_8$ & Is report formatting correct and consistent (e.g., valid Markdown heading syntax, renderable Markdown tables, consistent numbering, consistent emphasis/code styling, consistent citation format if used)? & Further specified and extended Q9 in LiveResearchBench's Table 3 \\
$p_9$ & Is the writing clear and professional at the sentence level (consistent tense/voice, minimal colloquialisms, avoids rhetorical exaggeration), with consistent terminology and abbreviation handling (define once, then reuse consistently)? & Further specified Q9 in LiveResearchBench's Table 3 and GRR48 in Dr.~Bench. Rewritten so that professionalism is defined more clearly.\\
$p_{10}$ & Are key terms, symbols, and abbreviations formatted consistently (e.g., italicization, capitalization, acronym, bolding), and is there no drifting where the same concept is labeled multiple ways without intent? & Newly written. Stylistic considerations are missing from previous evaluations, which is important for report presentation.\\
\bottomrule
\end{tabular}
\caption{\textbf{Presentation Evaluation Questions}}
\label{tab:presentation_questions}
\end{table}

%% file: latex/tables/human_studies_cov_judge.tex
\begin{table}[h]
\centering
\small
\begin{tabular}{lc}
\toprule
& \textbf{LLM-Human Agreement Rate} \\
\midrule
Turn 1 & 95.0\% \\
Turn 2 & 96.7\% \\
All & 95.8\% \\
\bottomrule
\end{tabular}
\caption{\textbf{LLM judge--human agreement rate across turns.}}
\label{tab:llm_human_agreement_cov}
\end{table}

%% file: latex/tables/format_seeds.tex
\begin{table*}[t]
\centering
\small
\begin{tabularx}{\textwidth}{cX}
\toprule
\textbf{ID} & \textbf{Feedback} \\
\midrule
1	&	Please rewrite this so the language is clearer and more straightforward, suitable for a reader with no prior knowledge.	\\
2	&	Whenever you introduce a technical concept, add a simple and real-world analogy to illustrate it.	\\
3	&	Standardize heading levels and naming so similar sections use parallel phrasing (e.g., ‘Approach’, ‘Results’, ‘Limitations’).	\\
4	&	Make sure that each section ends with a short summary sentence that emphasizes the main takeaway.	\\
5	&	Add a concise TL;DR at the beginning of the report that states the main question and key takeaways from the report.	\\
6	&	It would help if the report indicated which parts are essential reading and which parts are optional background.	\\
7	&	Highlight key sentences or phrases (e.g., with bold) so I can quickly find the most important takeaways.	\\
8	&	Please add short ‘section previews’ at the start of each main section, summarizing in 1–2 lines what will be covered.	\\
9	&	Please keep the core sections concise and move extended explanations, detailed justifications, and long background passages into clearly labeled ‘Appendix’ sections at the end.	\\
10	&	Consider adding transition sentences between sections to show how each part connects to the next.	\\
11	&	Add subheadings every 2-3 paragraphs to help readers navigate and find information quickly.	\\
12	&	Include a glossary of key terms at the end for readers who want quick reference.	\\
13	&	Consider using bullet points or numbered lists when presenting multiple related items rather than embedding them in prose.	\\
14	&	Add visual breaks like pull quotes to highlight critical insights so that it's easier to find takeaways.	\\
15	&	Apply bold formatting to critical findings, main conclusions, and essential terms on first mention, while using italics for secondary emphasis, technical terms in context, or when citing specific examples.	\\
16	&	Add a "How to Read This Report" section that explains the document's structure and what different readers should focus on.	\\
17	&	Vary sentence length and structure to maintain reader interest and create rhythm.	\\
18	&	Use "we" as much as possible than “you” or third-person pronouns to create connection with readers rather than maintaining complete detachment.	\\
19	&	Add a brief "Why This Matters" box at the start of technical sections to motivate readers.	\\
20	&	Close with actionable next steps or recommendations for related information so readers know what to do or read next.	\\
21	&	Create a separate "Frequently Asked Questions" section to address common points of confusion.	\\
\bottomrule
\end{tabularx}
\caption{21 Seed Format Feedback Pieces in \ours.}
\label{tab:format_seeds}
\end{table*}

%% file: latex/tables/human_studies.tex
\begin{table}[h]
\centering
\small
\begin{tabular}[width=\linewidth]{lcc}
\toprule
& \textbf{Content} & \textbf{Format} \\
\midrule
Naturalness & 100\% & 100\% \\
Draft-specificness & 92\% & 90\% \\
Actionability & 98\% & 98\% \\
Content-preserving & -- & 100\% \\
\midrule
Agreement Rate & 98\% & 96\% \\
\bottomrule
\end{tabular}
\caption{\textbf{Human verification results of simulated feedback quality.}}
\label{tab:feedback_quality}
\end{table}

%% file: latex/tables/citation_full.tex
\definecolor{turntwo}{gray}{0.96}
\providecommand{\sysname}[1]{\parbox[c]{1.1cm}{\raggedright #1}}
\newcommand{\sysTuluTight}{\sysname{DR\\Tulu}}

\begin{table*}[t]
\centering
\footnotesize
\setlength{\tabcolsep}{1.7pt}
\renewcommand{\arraystretch}{1.15}

\begin{adjustbox}{max width=.98\textwidth}
\begin{tabularx}{1.14\textwidth}{
    >{\raggedright\arraybackslash}p{1.1cm}
    >{\centering\arraybackslash}p{1.3cm}
    *{4}{>{\centering\arraybackslash}X}
    @{\hspace{3pt}}!{\vrule width 0.6pt}@{\hspace{3pt}}
    *{4}{>{\centering\arraybackslash}X}
    @{\hspace{3pt}}!{\vrule width 0.6pt}@{\hspace{3pt}}
    *{4}{>{\centering\arraybackslash}X}
    @{\hspace{3pt}}!{\vrule width 0.6pt}@{\hspace{3pt}}
    *{4}{>{\centering\arraybackslash}X}
}
\toprule
\multicolumn{2}{l}{} &
\multicolumn{4}{c}{\textbf{ResearchRubrics}} &
\multicolumn{4}{c}{\textbf{Dr.~Bench}} &
\multicolumn{4}{c}{\textbf{ResearcherBench}} &
\multicolumn{4}{c}{\textbf{Avg}} \\
\cmidrule(lr){3-6}\cmidrule(lr){7-10}\cmidrule(lr){11-14}\cmidrule(lr){15-18}
\textbf{Agent} & \textbf{Setting} &
\textbf{$|\mathcal{E}|$} & \textbf{$|\mathcal{E}_\text{cited}|$} & \textbf{$|\mathcal{S}|$} & \textbf{$|\mathcal{U}|$} &
\textbf{$|\mathcal{E}|$} & \textbf{$|\mathcal{E}_\text{cited}|$} & \textbf{$|\mathcal{S}|$} & \textbf{$|\mathcal{U}|$} &
\textbf{$|\mathcal{E}|$} & \textbf{$|\mathcal{E}_\text{cited}|$} & \textbf{$|\mathcal{S}|$} & \textbf{$|\mathcal{U}|$} &
\textbf{$|\mathcal{E}|$} & \textbf{$|\mathcal{E}_\text{cited}|$} & \textbf{$|\mathcal{S}|$} & \textbf{$|\mathcal{U}|$} \\
\midrule

 & Init & 73.3 & 29.5 & 20.4 & 15.1 & 73.8 & 36.1 & 23.7 & 19.1 & 44.8 & 22.2 & 17.7 & 11.4 & 64.0 & 29.3 & 20.6 & 15.2 \\
\rowcolor{turntwo}\cellcolor{white} & Reflect & \sgn{-2.8} & \sgn{-8.3} & \sgn{-6.8} & \sgn{-2.6} & \sgn{-2.4} & \sgn{-4.6} & \sgn{-3.8} & \sgn{-1.2} & \sgn{-0.5} & \sgn{-3.2} & \sgn{-2.8} & \sgn{-3.3} & \sgn{-1.9} & \sgn{-5.3} & \sgn{-4.5} & \sgn{-2.4} \\
\rowcolor{turntwo}\cellcolor{white} & Content$_1$ & \sgn{-7.2} & \sgn{-6.1} & \sgn{-8.3} & \sgn{-7.2} & \sgn{-11.6} & \sgn{-7.1} & \sgn{-11.3} & \sgn{-8.0} & \sgn{-6.8} & \sgn{-5.8} & \sgn{-7.9} & \sgn{-4.7} & \sgn{-8.5} & \sgn{-6.3} & \sgn{-9.2} & \sgn{-6.6} \\
\rowcolor{turntwo}\cellcolor{white}\multirow{-4}{*}{\sysOpenAI} & Format & \sgn{+2.4} & \sgn{-8.6} & \sgn{-9.6} & \sgn{-7.2} & \sgn{-4.9} & \sgn{-7.5} & \sgn{-8.0} & \sgn{-9.2} & \sgn{-0.9} & \sgn{-3.2} & \sgn{-4.4} & \sgn{-3.7} & \sgn{-1.1} & \sgn{-6.4} & \sgn{-7.3} & \sgn{-6.7} \\
\midrule

 & Init & 148.5 & 116.5 & 85.1 & 30.1 & 162.8 & 136.6 & 105.5 & 33.4 & 148.0 & 124.2 & 98.4 & 32.3 & 153.1 & 125.8 & 96.4 & 31.9 \\
\rowcolor{turntwo}\cellcolor{white} & Reflect & \sgn{-22.0} & \sgn{-95.2} & \sgn{-74.9} & \sgn{-21.5} & \sgn{-25.3} & \sgn{-107} & \sgn{-92.4} & \sgn{-22.6} & \sgn{-43.8} & \sgn{-106} & \sgn{-93.6} & \sgn{-25.6} & \sgn{-30.4} & \sgn{-101} & \sgn{-87.0} & \sgn{-23.2} \\
\rowcolor{turntwo}\cellcolor{white} & Content$_1$ & \sgn{+0.2} & \sgn{-1.0} & \sgn{-28.9} & \sgn{-3.2} & \sgn{-25.7} & \sgn{-31.6} & \sgn{-49.0} & \sgn{-7.5} & \sgn{+3.2} & \sgn{+9.6} & \sgn{-10.2} & \sgn{-0.3} & \sgn{-7.4} & \sgn{-7.7} & \sgn{-29.4} & \sgn{-3.6} \\
\rowcolor{turntwo}\cellcolor{white}\multirow{-4}{*}{\sysPerp} & Format & \sgn{-30.6} & \sgn{-35.0} & \sgn{-51.0} & \sgn{-9.8} & \sgn{-40.5} & \sgn{-46.7} & \sgn{-62.6} & \sgn{-8.6} & \sgn{-24.0} & \sgn{-26.6} & \sgn{-48.6} & \sgn{-4.7} & \sgn{-31.7} & \sgn{-36.1} & \sgn{-54.0} & \sgn{-7.7} \\
\midrule

 & Init & 64.3 & 33.7 & 24.6 & 16.3 & 61.4 & 36.5 & 26.8 & 19.1 & 60.4 & 37.0 & 30.4 & 19.6 & 62.0 & 35.7 & 27.3 & 18.3 \\
\rowcolor{turntwo}\cellcolor{white} & Reflect & \sgn{+10.5} & \sgn{+4.4} & \sgn{+1.2} & \sgn{+2.3} & \sgn{+11.6} & \sgn{+8.6} & \sgn{+5.4} & \sgn{+1.9} & \sgn{+9.6} & \sgn{+1.4} & \sgn{-1.7} & \sgn{-1.3} & \sgn{+10.6} & \sgn{+4.8} & \sgn{+1.7} & \sgn{+0.9} \\
\rowcolor{turntwo}\cellcolor{white} & Content$_1$ & \sgn{+1.8} & \sgn{-4.4} & \sgn{-4.3} & \sgn{-3.5} & \sgn{+0.2} & \sgn{-2.9} & \sgn{-3.8} & \sgn{-3.5} & \sgn{+2.7} & \sgn{-9.2} & \sgn{-10.2} & \sgn{-5.9} & \sgn{+1.6} & \sgn{-5.5} & \sgn{-6.1} & \sgn{-4.3} \\
\rowcolor{turntwo}\cellcolor{white}\multirow{-4}{*}{\sysOpen} & Format & \sgn{+1.6} & \sgn{-1.2} & \sgn{+0.3} & \sgn{-1.2} & \sgn{+3.2} & \sgn{-3.6} & \sgn{-3.2} & \sgn{-3.7} & \sgn{-0.1} & \sgn{-8.9} & \sgn{-9.0} & \sgn{-3.1} & \sgn{+1.6} & \sgn{-4.6} & \sgn{-3.9} & \sgn{-2.6} \\
\midrule

 & Init & 93.7 & 65.8 & 43.2 & 18.6 & 89.4 & 68.5 & 42.5 & 19.1 & 70.2 & 54.3 & 43.2 & 18.5 & 84.5 & 62.9 & 43.0 & 18.7 \\
\rowcolor{turntwo}\cellcolor{white} & Reflect & \sgn{-1.7} & \sgn{+0.9} & \sgn{-1.3} & \sgn{+0.0} & \sgn{+0.4} & \sgn{+1.6} & \sgn{+0.3} & \sgn{+0.4} & \sgn{-2.2} & \sgn{-2.5} & \sgn{-4.0} & \sgn{-1.7} & \sgn{-1.2} & \sgn{+0.0} & \sgn{-1.6} & \sgn{-0.4} \\
\rowcolor{turntwo}\cellcolor{white} & Content$_1$ & \sgn{+1.0} & \sgn{+2.9} & \sgn{-0.9} & \sgn{+2.7} & \sgn{+1.4} & \sgn{+3.9} & \sgn{-0.5} & \sgn{+5.3} & \sgn{+5.9} & \sgn{+1.9} & \sgn{-0.5} & \sgn{+3.2} & \sgn{+2.8} & \sgn{+2.9} & \sgn{-0.6} & \sgn{+3.8} \\
\rowcolor{turntwo}\cellcolor{white}\multirow{-4}{*}{\sysTuluTight} & Format & \sgn{-0.6} & \sgn{+3.9} & \sgn{+2.9} & \sgn{+0.4} & \sgn{+4.0} & \sgn{+6.3} & \sgn{+5.1} & \sgn{+2.1} & \sgn{-0.7} & \sgn{-0.4} & \sgn{-2.4} & \sgn{-1.0} & \sgn{+0.9} & \sgn{+3.3} & \sgn{+1.8} & \sgn{+0.5} \\

\bottomrule
\end{tabularx}
\end{adjustbox}

\caption{\textbf{Full Citation-related Results.} For each dataset, we report the average number of extracted claims ($|\mathcal{E}|$), claims with at least one citation ($|\mathcal{E}_\text{cited}|$), supported claims ($|\mathcal{S}|$), and citation counts ($|\mathcal{U}|$). Avg is the four counts averaged across all samples in three datasets.}
\label{tab:citation-stats}
\end{table*}

%% file: latex/figures/multi_turn_4x5_appendix.tex
\begin{figure*}[t]
    \centering
    \includegraphics[width=1\linewidth]{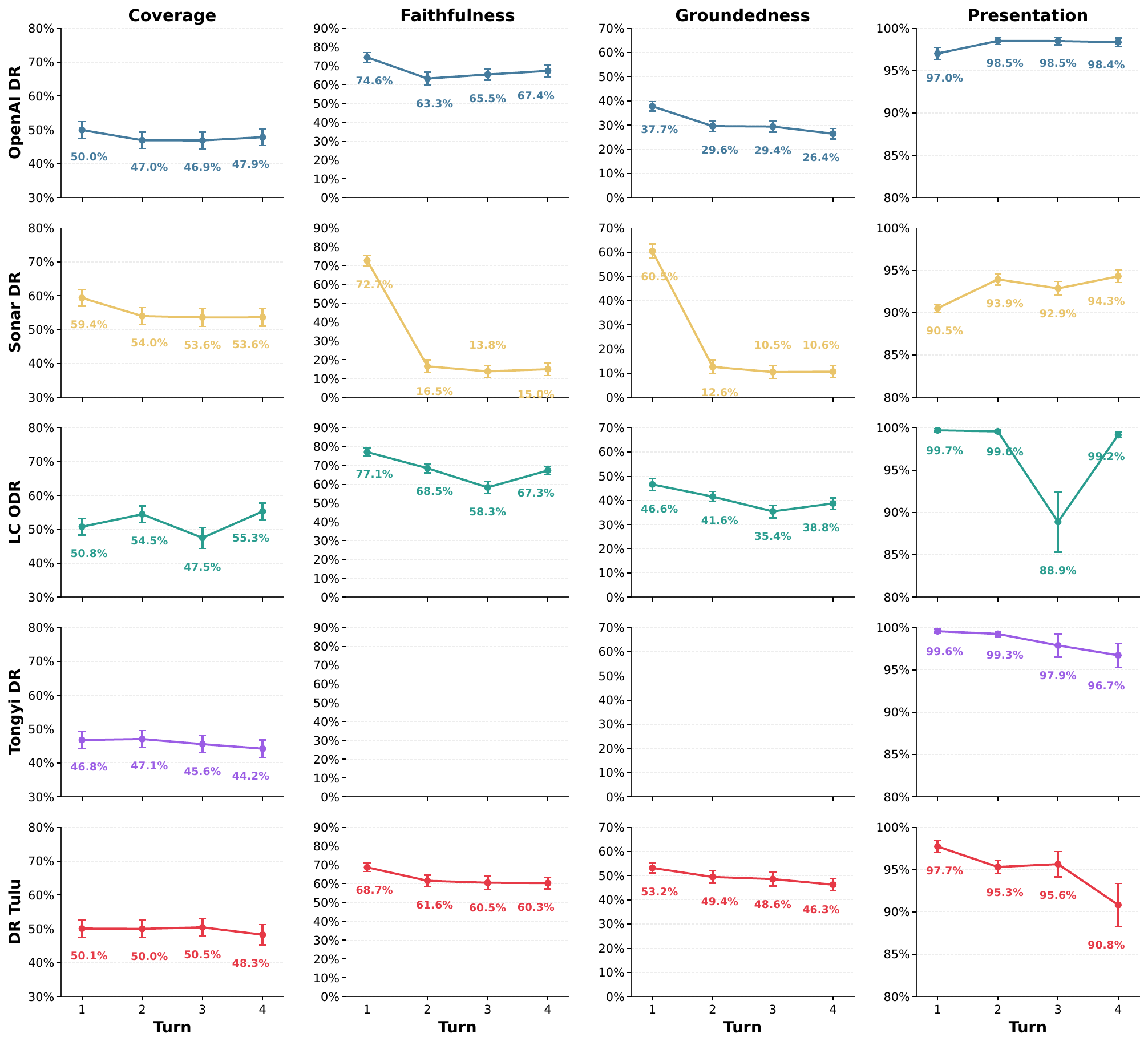}
    \caption{\textbf{Full multi-turn results for the Reflect setting.} Tongyi DR's citation faithfulness and claim groundedness are omitted since it is not trained to generate citations. Error bars indicate standard errors.}
    \label{fig:multi_turn_reflect}
\end{figure*}

\begin{figure*}[t]
    \centering
    \includegraphics[width=1\linewidth]{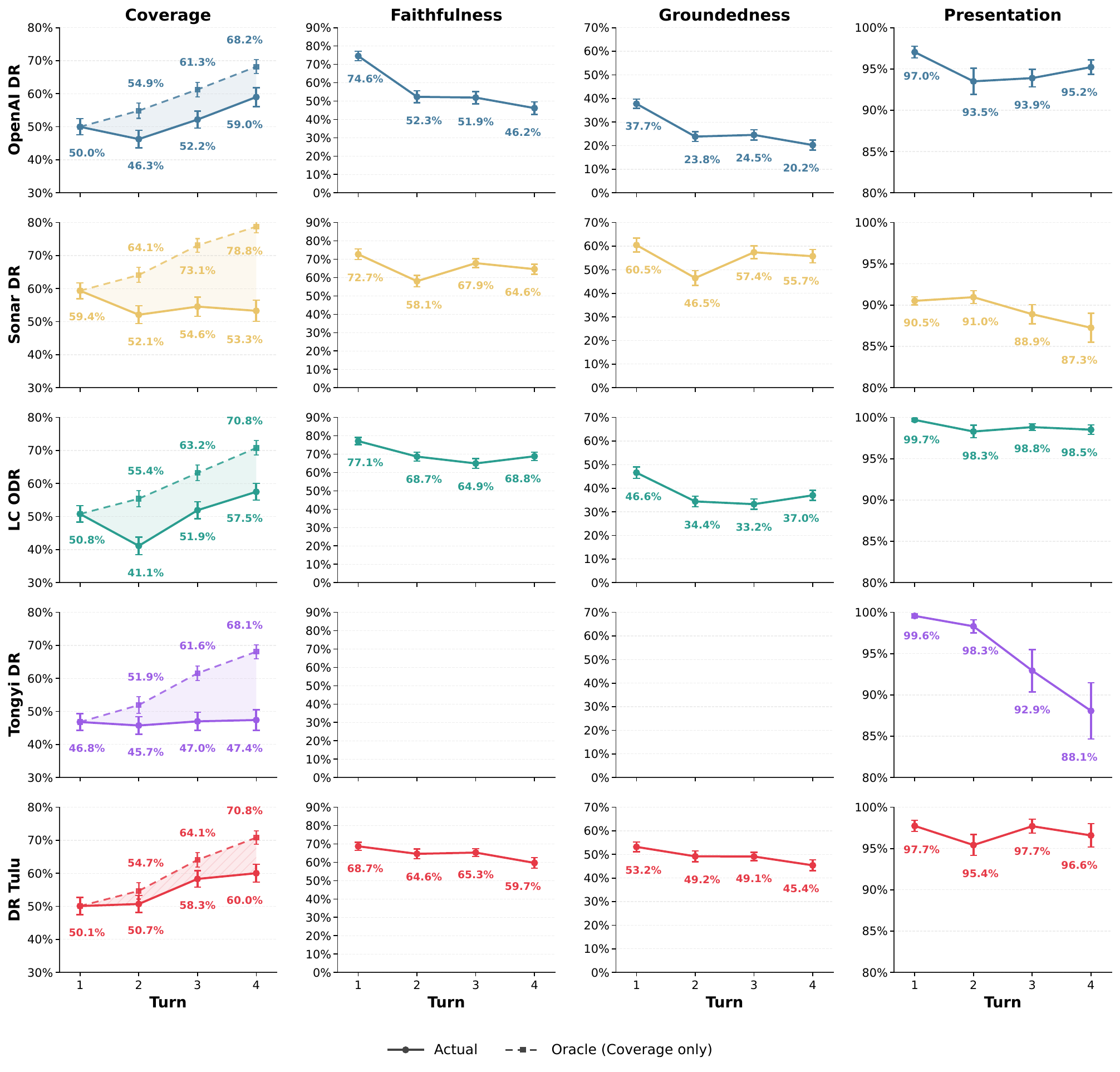}
    \caption{\textbf{Full multi-turn results for the Content$_1$ setting.} Tongyi DR's citation faithfulness and claim groundedness are omitted since it is not trained to generate citations. Error bars indicate standard errors.}
    \label{fig:multi_turn_content1}
\end{figure*}

%% file: latex/figures/multi-tiem_5x6.tex
\begin{figure*}[t]
    \centering
    \includegraphics[width=1\linewidth]{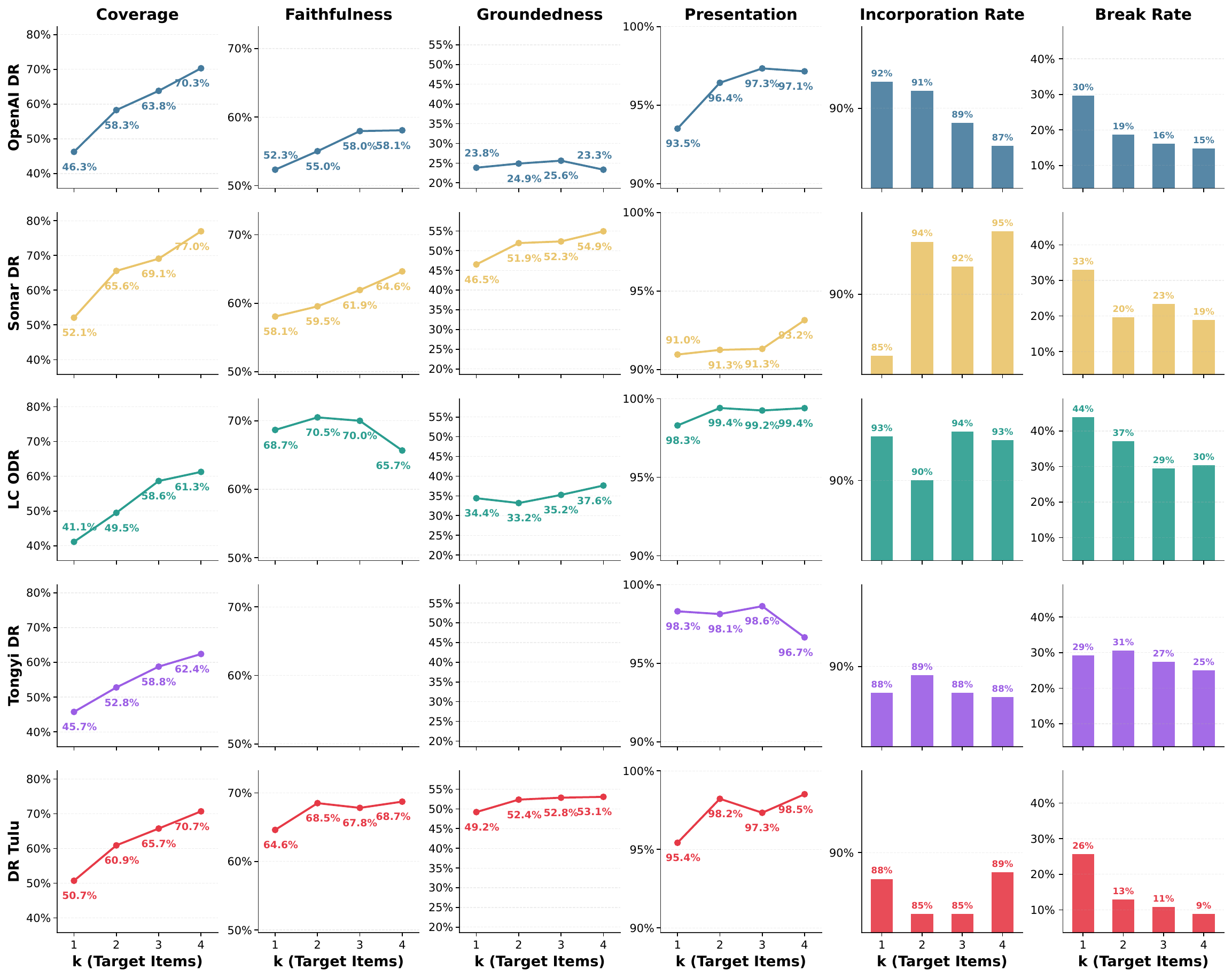}
    \caption{\textbf{Full results for content feedback with multiple feedback targets ($k$).}}
    \label{fig:multi_item_full}
\end{figure*}

%% file: latex/tables/o3-results.tex
\begin{table}[t]
\centering
\fontsize{7.5}{7.5}\selectfont
\setlength{\tabcolsep}{2.8pt}
\renewcommand{\arraystretch}{1.08}
\begin{adjustbox}{max width=\linewidth}
\begin{tabularx}{0.75\linewidth}{
    >{\raggedright\arraybackslash}p{1.1cm}
    *{4}{>{\centering\arraybackslash}X}
    @{\hspace{3pt}}!{\vrule width 0.6pt}@{\hspace{3pt}}
    *{2}{>{\centering\arraybackslash}X}
}
\toprule
&
\multicolumn{6}{c}{\textbf{ResearchRubrics (10 samples)}} \\
\cmidrule(lr){2-7}
\textbf{Setting} &
\textbf{Cov.} & \textbf{Fa.} & \textbf{Gr.} & \textbf{Pre.} & \textbf{Inc.} & \textbf{Brk.} \\
\midrule
Initial           & 76.0 & 59.0 & 24.1 & 98.9 & \multicolumn{1}{c}{--} & \multicolumn{1}{c}{--} \\
\cmidrule(lr){1-7}
Reflect           & \sgn{+0.5} & \sgn{-21.7} & \sgn{-11.5} & 0.0 & \multicolumn{1}{c}{--} & 4.6 \\
Content$_1$       & \sgn{-11.3} & \sgn{-4.6} & 0.0 & \sgn{-3.3} & 80.0 & 19.1 \\
Format            & \sgn{-2.4} & \sgn{-14.3} & \sgn{-9.7} & \sgn{+1.1} & 100.0 & 8.4 \\
\bottomrule
\end{tabularx}
\end{adjustbox}
\caption{\textbf{\texttt{o3-deep-research} results on the ResearchRubrics subset (10 samples).} Second turn's score change from Initial (turn 1) is shown for the four main metrics, along with incorporation (Inc.) and break (Brk.) rates.}
\label{tab:o3-results}
\end{table}

%% file: latex/tables/feedback_example.tex
\begin{table*}[t]
\centering
\small
\setlength{\tabcolsep}{8pt}
\renewcommand{\arraystretch}{1.15}
\begin{tabularx}{\textwidth}{@{}l X@{}}
\toprule
\textbf{Feedback type} & \textbf{Feedback example} \\
\midrule
Format &
Including a glossary of key terms at the end of the report would greatly benefit beginners by providing a quick reference to important concepts like forward propagation, backpropagation, and optimization methods, helping to reinforce understanding as they read through the material. \\
\addlinespace[2pt]
Format &
Consider applying bold formatting to key technical terms, main product features, and critical advantages when they first appear, while using italics for secondary details like specific APIs or model names; this will help readers quickly identify the most important information and improve overall readability. \\
\midrule
Content$_1$ &
Thanks for covering the MYC pathway---it’s a great start! To make the overview stronger, could you also include how NELF-E affects other important genes or pathways like BRCA1, RAD51, or its role in promoter-proximal pausing in HCC? That would really round out the explanation. \\
\addlinespace[2pt]
Content$_1$ &
Hey, could you add a part explaining how S1 shows that smaller, high-quality datasets can match the performance of much larger ones? Right now, it mostly talks about large-scale data but misses that important insight about data quality over quantity. \\
\addlinespace[2pt]
Content$_3$ &
Hey, the report would be way stronger if it included a clear marketing strategy covering at least four channels like social media influencers, local events, digital ads, and food delivery promos, with a quick note on how each helps build the brand. Also, it’s important to add staffing details for each concept---like specific roles needed for the food truck, fine dining, and fast casual spots---so we get a better sense of the team structure. Lastly, while naming a couple of design firms was helpful, including their contact info and some rough cost estimates would make it easier to move forward and compare options. \\
\bottomrule
\end{tabularx}
\caption{\textbf{Format and Content$_k$ feedback examples.} We show representative format feedback and content feedback with one or three targets.}
\label{tab:more_feedback_examples}
\end{table*}

%% file: latex/figures/examples/case_example.tex
\begin{figure}[t]
\begin{CaseBox}
\begin{lstlisting}[style=caseboxstyle]
BEFORE FEEDBACK:

1. DeepSeek V3: Advancing Architecture and Training Efficiency in Large Language Models
DeepSeek V3, an open-source LLM with 671 billion parameters but activating only 37 billion per inference, introduced significant architectural advances that have influenced LLM design...  @@[1][6][7][9][12][18][34]@@.
...
[1] Why DeepSeek v3 matters in the world of LLMs - Kiseki Labs: https://www.kisekilabs.com/blog-posts/why-deepseek-v3-matters-in-the-world-of-llms
[2] DeepSeek-R1 Release: https://api-docs.deepseek.com/news/news250120
...
@@[33]@@ AI: Deepest Lessons from DeepSeek. RTZ #615 - Substack: https://michaelparekh.substack.com/p/ai-deepest-lessons-from-deepseek
@@[34]@@ A Technical Tour of the DeepSeek Models from V3 to V3.2 - Sebastian Raschka Magazine: https://magazine.sebastianraschka.com/p/technical-deepseek
\end{lstlisting}
\tcbline
\begin{lstlisting}[style=caseboxstyle]
FEEDBACK:

The report could be improved by explicitly highlighting R1's standout creative task performance and diving deeper into the specialized domain optimization strategies it uses, as these details really show why it's a strong open-source alternative. Adding these points would give a clearer picture of its unique strengths beyond just benchmark scores.
\end{lstlisting}
\tcbline
\begin{lstlisting}[style=caseboxstyle]
AFTER FEEDBACK:

1. Architectural and Training Innovations
Mixture-of-Experts Architecture with Dynamic, Balanced Routing. DeepSeek R1 uses a massive Mixture-of-Experts (MoE) transformer structure hosting approximately 671 billion parameters... (*@\hl{([MISSING] in context citations)}@*)
...
[1] DeepSeek R1: Architectural Innovation in Reasoning Models: https://mbrenndoerfer.com/writing/deepseek-r1-architectural-innovation-reasoning-models
[2] DeepSeek-R1: Technical Overview of its Architecture and Innovations: https://www.geeksforgeeks.org/artificial-intelligence/
...
(*@\hl{[19]}@*) DeepSeek-R1: Features, o1 Comparison, Distilled Models: https://www.datacamp.com/blog/deepseek-r1
(*@\hl{[20]}@*) DeepSeek Reasoning for MLPerf Inference v5.1: https://mlcommons.org/2025/09/deepseek-inference-5-1/
\end{lstlisting}
\end{CaseBox}
 \caption{\textbf{Citation Degradation Example.} After revision, the model reduces the number of citations and omits in-context citations from the original response.}
\label{fig:example_prompt_missing_content}
\end{figure}

\begin{figure}
\centering
\begin{CaseBox}
\begin{lstlisting}[style=caseboxstyle]
BEFORE FEEDBACK:

A Brief Company Background: From Startup to Tech Titan:
Founded in 2003 by Peter Thiel, Nathan Gettings, Joe Lonsdale, Stephen Cohen, and Alex Karp, Palantir Technologies began life as a startup driven by a singular ambition: building sophisticated software capable of integrating, analyzing, and visualizing huge volumes of disparate data. Its name comes from the "seeing stones" in J.R.R. Tolkien's works---a nod to the company's mission of offering unparalleled insight through data fusion.
Initially, Palantir developed Gotham, a powerful platform customized for intelligence agencies, law enforcement, and national security organizations. Gotham enabled analysts to correlate mass surveillance, crime, and threat intelligence, dramatically increasing their ability to spot patterns invisible to traditional methods. Simultaneously, Palantir created Foundry for commercial clients, focusing on logistics, health care, manufacturing, and finance---providing companies the same analytical muscle as top-tier intelligence services.
@@From Secrecy to Public Markets: For years, Palantir operated largely behind closed doors, serving U.S. government entities including the CIA, FBI, and Department of Homeland Security. Only in September 2020 did the company go public via a direct listing, triggering a surge of interest from institutional investors and analysts seeking exposure to disruptive data technologies. While early revenue primarily stemmed from government contracts, the IPO marked Palantir's commitment to broaden its reach, signaling that its most valuable innovations would soon impact sectors far beyond security and intelligence.@@
\end{lstlisting}

\tcbline
\begin{lstlisting}[style=caseboxstyle]
FEEDBACK:

To enhance reader engagement throughout this detailed article, try varying sentence length and structure more deliberately---mix shorter, punchier sentences with longer, complex ones to create a natural rhythm that keeps the reader's attention from start to finish.
\end{lstlisting}
\tcbline
\begin{lstlisting}[style=caseboxstyle]
AFTER FEEDBACK:

Origins: From Startup Dreams to National Security Pillar
Founded in 2003 by Peter Thiel, Nathan Gettings, Joe Lonsdale, Stephen Cohen, and Alex Karp, Palantir began as an audacious bet on big data. Its name, inspired by Tolkien's "seeing stones", promised something radical: software that could stitch together massive, messy datasets into clear human insights.
Initially, Palantir courted governments hungry for answers post-9/11. Its Gotham platform became vital to CIA, FBI, and Department of Homeland Security analysts. It linked surveillance feeds, financial records, and travel logs-a digital Sherlock Holmes that spotted threats humans missed. For years, Palantir operated in shadow. Its first clients were classified. Its milestones unannounced. The IPO in September 2020 was its first public debut, yet the secret behind its power had only just begun to surface.
(*@\hl{([MISSING] paragraph)}@*)
\end{lstlisting}

\end{CaseBox}
\caption{\textbf{Example of missing content.} The revised response preserves the overall narrative but omits the final paragraph from the original.}
\label{fig:example_missing_facts}
\end{figure}

%% file: latex/figures/examples/content_feedback_prompt.tex
\begin{figure}
\begin{tcolorbox}[colback=ourSuperLightViolet,colframe=ourViolet,title=Content$_1$ feedback Simulation System Prompt]
\begin{VerbatimWrap}
You are a user providing feedback to a research report writing agent.

You will be provided with:
    1. The original query that you asked
    2. A specific evaluation rule that was used to assess the report where the agent achieves a suboptimal score
    3. The coverage status (whether the rule was covered in the report)
    4. The weight of the rule (positive means the rule should be covered, negative means the rule should NOT be covered)
    5. The evaluator's explanation for the score

Your task is to provide natural, constructive, concrete feedback that a normal user would give to improve the report based on this specific evaluation rule.

Guidelines:
    - Be as conversational and natural as possible. Imagine you are talking to a collaborator who is helping you to write the report.
    - Occasionally, you can use some colloquial language to make the feedback more realistic.
    - Your feedback should be in 1-2 sentences that is concise and to the point.
    - Don't repeat the evaluation explanation verbatim but use it as a reference to help you provide the feedback.
\end{VerbatimWrap}
\end{tcolorbox}
\caption{\textbf{Content$_1$ feedback Simulation System Prompt.}}
\label{fig:content1_sys_prompt}
\end{figure}

\begin{figure}
\begin{tcolorbox}[colback=ourSuperLightViolet,colframe=ourViolet,title=Content$_k$ feedback Simulation System Prompt]
\begin{VerbatimWrap}
You are a user providing feedback to a research report writing agent.

You will be provided with:
    1. The original query that you asked
    2. Several specific evaluation rules that were used to assess the report where the agent achieves suboptimal scores
    3. For each rule, the coverage status (whether the rule was covered in the report)
    4. For each rule, the weight of the rule (positive means the rule should be covered, negative means the rule should NOT be covered)
    5. The evaluator's explanation for each score

Your task is to provide natural, constructive, concrete feedback that a normal user would give to improve the report based on these specific evaluation rules.

Guidelines:
    - Be converstational and natural as possible. Imagine you are talking to a collaborator who is helping you to write the report.
    - Occasionally, you can use some colloquial language to make the feedback more realistic.
    - Your feedback for each evaluation point should be in 1-2 sentences that is concise and to the point.
    - Don't repeat the evaluation explanations verbatim but use them as a reference to help you provide the feedback.
\end{VerbatimWrap}
\end{tcolorbox}
\caption{\textbf{Content$_k$ feedback Simulation System Prompt.}}
\label{fig:contentk_sys_prompt}
\end{figure}

\begin{figure}
\begin{tcolorbox}[colback=ourSuperLightViolet,colframe=ourViolet,title=Format feedback Simulation System Prompt]
\begin{VerbatimWrap}
You are a user providing feedback on the report's writing, structure, and presentation only, not on its facts, reasoning, or conclusions.

You will be provided with:
1. The original query given to the agent
2. The agent-generated research report
3. Three seed feedback examples from a predefined list

Your task is to:
1. First, select which of the three feedback examples would be most suitable and relevant for improving this report. It should be targeting an aspect that the report misses or did not do well. Do not give feedback to what is already done in the report. For example, if the report already uses subheadings or bulleted lists to organize information, you should not give feedback on that but select another aspect to ask for improvement.
2. Then, start from the selected feedback and either (a) rewrite it to be more specific and tailored to the actual content of the report while preserving its core suggestion, or (b) if a slightly different but closely related suggestion would better improve this particular report, adapt it into that alternative while staying within the same improvement category.

Your final feedback must adhere to the following specific desiderata:
- **Content-preserving**: Your feedback must not require any edits to existing content in the current draft. It should only incur changes in the form, structure, organization, tone, or style of the writing. Do NOT ask for new evidence, new arguments, or different conclusions.
- **Naturalness**: The language and wording should be natural and human-like, as if it was a natural follow-up response from the user themselves, or a thoughtful peer/supervisor. Give exactly one coherent suggestion (1–2 sentences) that feels like a natural follow-up from the user.
- **Draft-specific**: The feedback should be tailored to the original query and the current draft of the report, targeting aspects that the current draft misses and have clear room for improvement.
- **Actionability**: The feedback should be concrete and actionable, phrasing as implementable suggestions and avoiding vague comments such as "improve clarity" without explaining how. Make it specific to this draft and clearly implementable.

Please only respond with the final rewritten feedback, without any additional explanation or commentary.
\end{VerbatimWrap}
\end{tcolorbox}
\caption{\textbf{Format feedback Simulation System Prompt.}}
\label{fig:format_sys_prompt}
\end{figure}

%% file: latex/figures/examples/evaluation_prompt.tex
\begin{figure}
\begin{tcolorbox}[colback=ourSuperLightViolet,colframe=ourViolet,title=Coverage Evaluation System Prompt]
\begin{VerbatimWrap}
You will be given a question the user asked (in <question></question> tags) and the corresponding report (in <report></report> tags) given as a response to the question by an assistant. You will then be given a specific criterion to evaluate the report against (in <criterion></criterion> tags. It could be a yes/no question or a statement about the report that you should judge whether it's true or not).

Your task is to score the report based on whether it satisfies the criterion or not on a three-point scale: 1.0 if the report satisfies the criterion, 0.5 if the report partially satisfies the criterion, 0.0 if the report does not satisfy the criterion. Judge only the specified aspect(s), not any other qualities of the report. Please also provide a short (2-3 sentences maximum) justification for your score. Note: A criterion might be positive or negative. Satisfying the criterion means that the report contains the content that is described by the criterion, which should not be confused with satisfying the user's request.

Output only a JSON string with the following format: {\"score\": float, \"justification\": string}. Do not include any other text or comments in your response.
\end{VerbatimWrap}
\end{tcolorbox}
\caption{\textbf{Coverage Evaluation System Prompt.}}
\label{fig:checklist_eval}
\end{figure}

\begin{figure}
\begin{tcolorbox}[colback=ourSuperLightViolet,colframe=ourViolet,title=Coverage Evaluation User Prompt]
\begin{VerbatimWrap}
Evaluate the report based on the given criterion.\n
<question>\n{question}\n</question>\n
<report>\n{report}\n</report>\n
<criterion>\n{criterion}\n</criterion>\n
\end{VerbatimWrap}
\end{tcolorbox}
\caption{\textbf{Coverage Evaluation User Prompt.}}
\label{fig:coverage_eval_user}
\end{figure}

\begin{figure}
\begin{tcolorbox}[colback=ourSuperLightViolet,colframe=ourViolet,title=Presentation Evaluation System Prompt]
\begin{VerbatimWrap}
You will be given a question the user asked (in <question></question> tags) and the corresponding report (in <report></report> tags) given as a response to the question by an assistant. You will then be given a specific criterion to evaluate the report against (in <criterion></criterion> tags).

Your task is to score the report based on whether it satisfies the criterion or not: 1 if the report satisfies the criterion and 0 if the report does not satisfy the criterion. You might also be asked to give score=-1 when the criterion is not applicable to the report. Please do that when instructed. Judge only the specified aspect(s) in the criterion, not any other qualities of the report. Please also provide a short (2 sentences maximum) justification for your score.

Output only a JSON string with the following format: {\"score\": int, \"justification\": string}. Do not include any other text or comments in your response.
\end{VerbatimWrap}
\end{tcolorbox}
\caption{\textbf{Presentation Evaluation System Prompt.}}
\label{fig:presentation_eval_sys}
\end{figure}

\begin{figure}
\begin{tcolorbox}[colback=ourSuperLightViolet,colframe=ourViolet,title=Presentation Evaluation User Prompt]
\begin{VerbatimWrap}
Evaluate the report based on the given criterion. If the criterion is not applicable to the report, score should be -1 instead of 0/1.
<question>\n{question}\n</question>\n
<report>\n{report}\n</report>\n
<criterion>\n{criterion}\n</criterion>
\end{VerbatimWrap}
\end{tcolorbox}
\caption{\textbf{Presentation Evaluation User Prompt.}}
\label{fig:presentation_eval_user}
\end{figure}

\begin{figure}
\begin{tcolorbox}[colback=ourSuperLightViolet,colframe=ourViolet,title=Claim Extraction System Prompt]
\begin{VerbatimWrap}
You will be provided with a research report (in <report></report> tags). The body of the report will contain many factual claims and citations to references. A section of the report will be highlighted in <highlighted_section></highlighted_section> tags.

Your task is to extract all factual claims from and only from this highlighted section, along with the corresponding citation URLs if they exist.

Extraction Guidelines:
- You should ONLY extract claims from the highlighted section. Other parts of the report should only be used as context.
- Each of these claims should be verifiable against external sources (e.g., via Wikipedia). Any story, personal experiences, hypotheticals (e.g.,\"would be\" or subjunctive), subjective statements (e.g., opinions), suggestions, advice, instructions, and other such content should not be included in the list.
- All extracted claims should be standalone that can be understandable and verifiable without additional context.
- You should preserve the original wording where possible, but provide necessary context to make the claim self-contained. Particularly, use the context to recover pronouns, anaphoric references (e.g. \"the paper\", \"the idea\"), and other such information to make the claim self-contained. Use the name of entities rather than anaphors whenever possible.
- Along with the claims, you should also extract the corresponding citation URL(s) if they exist. Citations can be in different formats:
- A segment of text + [number], for example: \"Li Qiang constructed a socioeconomic status index (SES) based on income, education, and occupation, dividing society into 7 levels [15]\"
- A segment of text + [number†(some line numbers, etc.)], for example: \"Li Qiang constructed a socioeconomic status index (SES) based on income, education, and occupation, dividing society into 7 levels [15†L10][5L23][7†summary]\"
- [Citation Source](Citation Link), for example: \"Bolsonaro's rhetoric and frequent conflicting signals (e.g. encouraging gatherings) eroded public trust in institutions [pmc.ncbi.nlm.nih.gov](https://pmc.ncbi.nlm.nih.gov/articles/PMC11042250/#:~:text=Conclusion).\"
If the citation format is among the first two, please refer to the references/sources section at the end to find the corresponding URLs for each claim.
- If a claim has no corresponding citation to support it, return an empty list for the url field.
"- If multiple claims are associated with the same citation, extract them as separate entries. If a claim has multiple citations, include all citation URLs in the url list.

Output format:
Return a list of JSON objects of the following format: [{\"claim\": \"EXTRACTED CLAIM TEXT\", \"url\": [\"URL1\", \"URL2\", ...]}, ...].
Output only the JSON list directly, without any chitchat or explanations. If the highlighted section does not contain any verifiable factual claims, please return an empty list. Please make sure the URLs are copied verbatim from the original citations. The \"url\" field should be a emtpy, single-item, or multi-item list.
\end{VerbatimWrap}
\end{tcolorbox}
\caption{\textbf{Claim Extraction System Prompt.}}
\label{fig:claim_extraction_sys_prompt}
\end{figure}

\begin{figure}
\begin{tcolorbox}[colback=ourSuperLightViolet,colframe=ourViolet,title=Claim Extraction User Prompt]
\begin{VerbatimWrap}
Extract the verifiable factual claims from the highlighted section of the report.\n
<report>\n{report}\n</report>\n
<highlighted_section>\n{highlighted_section}\n</highlighted_section>
\end{VerbatimWrap}
\end{tcolorbox}
\caption{\textbf{Claim Extraction User Prompt.}}
\label{fig:claim_extraction_user_prompt}
\end{figure}

\begin{figure}
\begin{tcolorbox}[colback=ourSuperLightViolet,colframe=ourViolet,title=Claim Supportedness Judge System Prompt]
\begin{VerbatimWrap}
You will be provided with a reference content (in <reference_content></reference_content> tags) and a claim or statement (in <claim></claim> tags). Your task is to determine whether the claim is 'supported', 'insufficient', or 'contradictory' with respect to the reference. Please note:

- 'supported': the claim is clearly supported by the reference.
- 'insufficient': the claim is weakly supported by the reference, or the reference is missing key evidence, or the claim is not related to the reference.
- 'contradictory': the claim contradicts the reference.
First, assess whether the reference contains any valid content. If the reference contains no valid information, such as a 'page not found' message, then the claim should be considered 'insufficient'. Then, carefully read the reference and the claim, and determine the relationship between the claim and the reference. The reference content can be from one or multiple webpages.

Output Format: Return a JSON string with the following format: {\"result\": \"supported\" | \"insufficient\" | \"contradictory\"}. Do not include any other text or comments in your response. Please make sure the result is based purely on whether the claim is supported by the reference, not any other factors.
\end{VerbatimWrap}
\end{tcolorbox}
\caption{\textbf{Claim Supportedness Judge System Prompt}}
\label{fig:supported_judge_sys_prompt}
\end{figure}

\begin{figure}
\begin{tcolorbox}[colback=ourSuperLightViolet,colframe=ourViolet,title=Claim Supportedness Judge User Prompt]
\begin{VerbatimWrap}
Judge if the cited reference content supports the claim.\n
<reference_content>\n{url_content}\n</reference_content>\n
<claim>\n{claim}\n</claim>
\end{VerbatimWrap}
\end{tcolorbox}
\caption{\textbf{Claim Supportedness Judge User Prompt.}}
\label{fig:supported_judge_user_prompt}
\end{figure}

\begin{figure}
\begin{tcolorbox}[colback=ourSuperLightViolet,colframe=ourViolet,title=URL Content Summarization System Prompt]
\begin{VerbatimWrap}
You are a webpage summarization assistant. Your goal is to create a summary that preserves the most important information from the original web page. Given scraped webpage in markdown format (in <webpage_content></webpage_content> tags) and a list of claims (in <claims></claims> tags),
extract and summarize the parts of the webpage that are relevant to the claims.

Make sure you include all information that could support, contradict, or provide context for the claims. Also, preserve as much other key information in the webpage as possible to provide comprehensive context and be self-contained.

Try to use the original wording of the webpage content as much as possible.
If you find the webpage content is irrelevant to the claims, just generally summarize the web page content covering all key information.
When you are summarizing, DO NOT use the third-person perspective (e.g. the webpage states that ..., the author says that ..., etc.). Just consider you are shortening the webpage as the author.

Be as objective as possible and do not make any judgement or comments on the content.

Aim for about 20 percent of the original length, unless the webpage is already concise.
\end{VerbatimWrap}
\end{tcolorbox}
\caption{\textbf{URL Content Summarization System Prompt}}
\label{fig:url_summary_sys}
\end{figure}

\begin{figure}
\begin{tcolorbox}[colback=ourSuperLightViolet,colframe=ourViolet,title=URL Content Summarization User Prompt.]
\begin{VerbatimWrap}
Summarize the webpage content that are potentially relevant to the claims.\n
<webpage_content>\n{content}\n</webpage_content>\n
<claims>\n{claims}\n</claims>\n
Provide a summary of the webpage content. Preserve the original wording of the webpage content as much as possible, and include all meaningful details. Do not include any other text or explanations in your response.
\end{VerbatimWrap}
\end{tcolorbox}
\caption{\textbf{URL Content Summarization User Prompt.}}
\label{fig:url_summary_user}
\end{figure}

\begin{figure}
\begin{tcolorbox}[colback=ourSuperLightViolet,colframe=ourViolet,title=Pairwise Formal Feedback Judge System Prompt]
\begin{VerbatimWrap}
You will be given a research question the user asked (in <question></question> tags) and two versions of the report, original (in <report></report> tags) and revised (in <revised_report></revised_report> tags) that are generated by an assistant. The revised report is a revised version of the original report based on the feedback (in <feedback></feedback> tags) provided by the user.

Your task is to score the revised report based on whether it incorporates the feedback provided by the user or not, comparing it to the original report: 1.0 if the revised report incorporates the feedback, 0.5 if the revised report partially incorporates the feedback, and 0.0 if the revised report does not incorporate the feedback.

Output only a JSON string with the following format: {\"score\": float}. Do not include any other text or comments in your response. Please make sure the score is based purely on whether the feedback is reflected in the revised report compared to the original report, not any other factors.
\end{VerbatimWrap}
\end{tcolorbox}
\caption{\textbf{Pairwise Formal Feedback Judge System Prompt.}}
\label{fig:pair_judge_sys}
\end{figure}

\begin{figure}
\begin{tcolorbox}[colback=ourSuperLightViolet,colframe=ourViolet,title=Pairwise Formal Feedback Judge User Prompt]
\begin{VerbatimWrap}
Score the revised report based on whether it incorporates the feedback provided by the user compared to the original report.
<question>\n{question}\n</question>\n
<report>\n{report}\n</report>\n"
<revised_report>\n{revised_report}\n</revised_report>\n
<feedback>\n{feedback}\n</feedback>
\end{VerbatimWrap}
\end{tcolorbox}
\caption{\textbf{Pairwise Formal Feedback Judge User Prompt.}}
\label{fig:pair_judge}
\end{figure}

%% file: latex/figures/examples/refine_system_prompt.tex
\begin{figure}[t]
\begin{tcolorbox}[colback=ourSuperLightViolet,colframe=ourViolet,title=Prompt Engineering System Prompt]
\begin{VerbatimWrap}
You are an expert technical editor. Your task is to translate high-level user feedback into a minimal, localized edit plan.

Input You Will Receive:
1) Original Query
2) Full Research Report
3) Original Feedback (often vague or high-level)

Your Goal:
Create a structured **edit plan** that enables a research agent to take specific, localized editing actions without ambiguity.
Do NOT write or fabricate final factual content — your job is to **identify where** and **what type** of content needs to be added/changed.

---

Editing Constraints:

- Only use the following atomic actions: **DELETE / INSERT / MODIFY**
- Every edit must specify an exact location with:
  - `Section` name (must match exactly)
  - `Subsection` name (or N/A if not applicable)
  - `Anchor` quote: A short (≤18 words) **verbatim** sentence/phrase from the current report that clearly identifies **where** the edit should occur.
- Reference the Anchor in your **Content Spec** to clarify where in the text the change happens.
- INSERT actions can create new sections/subsections, but only if explicitly specified in the feedback.
- Do NOT invent specific facts (names, numbers, dates, benchmarks, claims).

---

Output Format (Markdown):

Feedback:
[Insert original feedback exactly as received]

Edit Actions:
1) Action: DELETE | INSERT | MODIFY
   Location:
   - Section: "[Exact section name]"
     *(For new sections, use format: `NEW: [Section Name]`)*
   - Subsection: "[Exact subsection name]" (or N/A)
     *(For new subsections, use format: `NEW: [Subsection Name]`)*
   - Anchor: "[Short verbatim quote from report]"
     *(For new sections/subsections, specify relative location, e.g., "After section 'Discussion')*
   Content Spec:
   - What to change: Describe required content, not final prose
   - Must-include: Specific elements that must be part of the edit
\end{VerbatimWrap}
\end{tcolorbox}
\caption{\textbf{Prompt Engineering System Prompt.}}
\label{fig:prompt_engi}
\end{figure}

%% file: latex/figures/examples/refiner_suffix.tex
\begin{figure}[t]
\begin{tcolorbox}[colback=ourSuperLightViolet,colframe=ourViolet,title=Prompt Engineering Hard-coded Constraint Suffix]
\begin{VerbatimWrap}
You are given:
1) The original user feedback
2) A structured list of localized Edit Actions derived from that feedback

Each Edit Action includes:
- Action: One of DELETE, INSERT, or MODIFY — the atomic type of edit to apply.
- Section / Subsection: The precise location in the document where the edit applies. If the action introduces a new section/subsection, it will be labeled as `NEW: [Name]`.
- Anchor: A short verbatim quote from the report identifying the exact insertion/modification point. For new sections/subsections, this is a relative reference (e.g., "After section 'Discussion'").
- Content Spec: A short explanation of what to change, localized to the Anchor location. This is NOT final content — only a structural and intent-level guide.

Non-negotiable editing constraints:
- Apply ONLY the actions listed under "Edit Actions".
- Do NOT infer, add, or modify edits beyond what is explicitly specified.
- Do NOT reinterpret or expand the original feedback.
- Do NOT rewrite sections wholesale; keep edits strictly local to the specified Anchor quote.

Your Task:
Apply the Edit Actions to improve the report by making precise, localized edits at the specified locations, adhering strictly to all constraints above. Please only output the revised report and no other text such as comments or explanations.
\end{VerbatimWrap}
\end{tcolorbox}
\caption{\textbf{Prompt Engineering Hard-coded Constraint Suffix.}}

\label{fig:refiner_suffix}
\end{figure}

%% file: latex/figures/examples/reviser_prompt.tex
\newpage
\begin{figure}[t]
\begin{tcolorbox}[colback=ourSuperLightViolet,colframe=ourViolet,title=Reviser Subagent System Prompt]
\begin{VerbatimWrap}
You are a research report revision assistant. Your task is to revise and improve a research report based on user feedback.

You have access to a web search tool to find additional information, verify facts, or gather supporting evidence. Note that you can only use the web search tool for {max_tool_calls} times. If you have used the web search tool for {max_tool_calls} times, you should then produce the final report and stop using the web search tool again.

## Guidelines

1. **Understand the feedback**: Carefully read the user's feedback to understand what needs to be improved.
2. **Search when needed**: If the user feedback requires searching for additional information, use the web search tool to find the information.
3. **Maintain Quality**: Ensure the revised report locally addresses the user feedback without making any changes to other parts.
\end{VerbatimWrap}
\end{tcolorbox}
\caption{\textbf{Reviser Subagent System Prompt.}}
\label{fig:reviser_sys}
\end{figure}

\begin{figure}[t]
\begin{tcolorbox}[colback=ourSuperLightViolet,colframe=ourViolet,title=Reviser Subagent User Prompt]
\begin{VerbatimWrap}
## Original Research Question

{question}

## Current Report

{report}

## Feedback for Revision

{feedback}

---

Please revise the report based on the feedback above. Use web search if you need additional information. Return ONLY the revised report and no other text such as comments or explanations.

\end{VerbatimWrap}
\end{tcolorbox}
\caption{\textbf{Reviser Subagent User Prompt.}}
\label{fig:reviser_user}
\end{figure}

%% file: main.bbl
\begin{thebibliography}{44}
\providecommand{\natexlab}[1]{#1}
\providecommand{\url}[1]{\texttt{#1}}
\expandafter\ifx\csname urlstyle\endcsname\relax
  \providecommand{\doi}[1]{doi: #1}\else
  \providecommand{\doi}{doi: \begingroup \urlstyle{rm}\Url}\fi

\bibitem[Arora et~al.(2025)Arora, Wei, Hicks, Bowman, Quiñonero-Candela,
  Tsimpourlas, Sharman, Shah, Vallone, Beutel, Heidecke, and
  Singhal]{arora2025healthbench}
Rahul~K. Arora, Jason Wei, Rebecca~Soskin Hicks, Preston Bowman, Joaquin
  Quiñonero-Candela, Foivos Tsimpourlas, Michael Sharman, Meghan Shah, Andrea
  Vallone, Alex Beutel, Johannes Heidecke, and Karan Singhal.
\newblock Healthbench: Evaluating large language models towards improved human
  health, 2025.
\newblock URL \url{https://arxiv.org/abs/2505.08775}.

\bibitem[Chen et~al.(2025)Chen, Ma, Zhuang, Nie, Zou, Liu, Green, Patel, Meng,
  Su, Sharifymoghaddam, Li, Hong, Shi, Liu, Thakur, Zhang, Gao, Chen, and
  Lin]{chen2025BrowseCompPlus}
Zijian Chen, Xueguang Ma, Shengyao Zhuang, Ping Nie, Kai Zou, Andrew Liu,
  Joshua Green, Kshama Patel, Ruoxi Meng, Mingyi Su, Sahel Sharifymoghaddam,
  Yanxi Li, Haoran Hong, Xinyu Shi, Xuye Liu, Nandan Thakur, Crystina Zhang,
  Luyu Gao, Wenhu Chen, and Jimmy Lin.
\newblock Browsecomp-plus: A more fair and transparent evaluation benchmark of
  deep-research agent, 2025.
\newblock URL \url{https://arxiv.org/abs/2508.06600}.

\bibitem[Du et~al.(2025)Du, Xu, Zhu, Wang, and Mao]{du2025deepresearch}
Mingxuan Du, Benfeng Xu, Chiwei Zhu, Xiaorui Wang, and Zhendong Mao.
\newblock Deepresearch bench: A comprehensive benchmark for deep research
  agents, 2025.
\newblock URL \url{https://arxiv.org/abs/2506.11763}.

\bibitem[Fan et~al.(2019)Fan, Jernite, Perez, Grangier, Weston, and
  Auli]{fan2019eli5}
Angela Fan, Yacine Jernite, Ethan Perez, David Grangier, Jason Weston, and
  Michael Auli.
\newblock {ELI}5: Long form question answering.
\newblock In Anna Korhonen, David Traum, and Llu{\'i}s M{\`a}rquez, editors,
  \emph{Proceedings of the 57th Annual Meeting of the Association for
  Computational Linguistics}, pages 3558--3567, Florence, Italy, July 2019.
  Association for Computational Linguistics.
\newblock \doi{10.18653/v1/P19-1346}.
\newblock URL \url{https://aclanthology.org/P19-1346/}.

\bibitem[Fan et~al.(2025)Fan, Niu, Zheng, Zhang, Huang, Chen, Lin, and
  Huang]{fan2025understandingdeepresearchreports}
Tianyu Fan, Xinyao Niu, Yuxiang Zheng, Fengji Zhang, Chengen Huang, Bei Chen,
  Junyang Lin, and Chao Huang.
\newblock Understanding deepresearch via reports, 2025.
\newblock URL \url{https://arxiv.org/abs/2510.07861}.

\bibitem[Gao et~al.(2023)Gao, Yen, Yu, and Chen]{gao2023alce}
Tianyu Gao, Howard Yen, Jiatong Yu, and Danqi Chen.
\newblock Enabling large language models to generate text with citations.
\newblock In Houda Bouamor, Juan Pino, and Kalika Bali, editors,
  \emph{Proceedings of the 2023 Conference on Empirical Methods in Natural
  Language Processing}, pages 6465--6488, Singapore, December 2023. Association
  for Computational Linguistics.
\newblock \doi{10.18653/v1/2023.emnlp-main.398}.
\newblock URL \url{https://aclanthology.org/2023.emnlp-main.398/}.

\bibitem[Gou et~al.(2024)Gou, Shao, Gong, Shen, Yang, Duan, and
  Chen]{gou2023critic}
Zhibin Gou, Zhihong Shao, Yeyun Gong, Yelong Shen, Yujiu Yang, Nan Duan, and
  Weizhu Chen.
\newblock Critic: Large language models can self-correct with tool-interactive
  critiquing, 2024.
\newblock URL \url{https://arxiv.org/abs/2305.11738}.

\bibitem[Han et~al.(2024)Han, Zhang, Qi, Xu, Wang, Liu, Wang, Min, and
  Castelli]{han2024rag}
Rujun Han, Yuhao Zhang, Peng Qi, Yumo Xu, Jenyuan Wang, Lan Liu, William~Yang
  Wang, Bonan Min, and Vittorio Castelli.
\newblock {RAG}-{QA} arena: Evaluating domain robustness for long-form
  retrieval augmented question answering.
\newblock In Yaser Al-Onaizan, Mohit Bansal, and Yun-Nung Chen, editors,
  \emph{Proceedings of the 2024 Conference on Empirical Methods in Natural
  Language Processing}, pages 4354--4374, Miami, Florida, USA, November 2024.
  Association for Computational Linguistics.
\newblock \doi{10.18653/v1/2024.emnlp-main.249}.
\newblock URL \url{https://aclanthology.org/2024.emnlp-main.249/}.

\bibitem[Han et~al.(2025)Han, Chen, CuiZhu, Miculicich, Sun, Bi, Wen, Wan, Wen,
  Ma{\^\i}tre, et~al.]{han2025deep}
Rujun Han, Yanfei Chen, Zoey CuiZhu, Lesly Miculicich, Guan Sun, Yuanjun Bi,
  Weiming Wen, Hui Wan, Chunfeng Wen, Sol{\`e}ne Ma{\^\i}tre, et~al.
\newblock Deep researcher with test-time diffusion.
\newblock \emph{arXiv preprint arXiv:2507.16075}, 2025.

\bibitem[Hashemi et~al.(2024)Hashemi, Eisner, Rosset, Van~Durme, and
  Kedzie]{hashemi2024llmrubrics}
Helia Hashemi, Jason Eisner, Corby Rosset, Benjamin Van~Durme, and Chris
  Kedzie.
\newblock {LLM}-rubric: A multidimensional, calibrated approach to automated
  evaluation of natural language texts.
\newblock In Lun-Wei Ku, Andre Martins, and Vivek Srikumar, editors,
  \emph{Proceedings of the 62nd Annual Meeting of the Association for
  Computational Linguistics (Volume 1: Long Papers)}, pages 13806--13834,
  Bangkok, Thailand, August 2024. Association for Computational Linguistics.
\newblock \doi{10.18653/v1/2024.acl-long.745}.
\newblock URL \url{https://aclanthology.org/2024.acl-long.745/}.

\bibitem[Huang et~al.(2024)Huang, Chen, Mishra, Zheng, Yu, Song, and
  Zhou]{huang2023large}
Jie Huang, Xinyun Chen, Swaroop Mishra, Huaixiu~Steven Zheng, Adams~Wei Yu,
  Xinying Song, and Denny Zhou.
\newblock Large language models cannot self-correct reasoning yet, 2024.
\newblock URL \url{https://arxiv.org/abs/2310.01798}.

\bibitem[Jiang et~al.(2023)Jiang, Xu, Gao, Sun, Liu, Dwivedi-Yu, Yang, Callan,
  and Neubig]{jiang2023active}
Zhengbao Jiang, Frank Xu, Luyu Gao, Zhiqing Sun, Qian Liu, Jane Dwivedi-Yu,
  Yiming Yang, Jamie Callan, and Graham Neubig.
\newblock Active retrieval augmented generation.
\newblock In Houda Bouamor, Juan Pino, and Kalika Bali, editors,
  \emph{Proceedings of the 2023 Conference on Empirical Methods in Natural
  Language Processing}, pages 7969--7992, Singapore, December 2023. Association
  for Computational Linguistics.
\newblock \doi{10.18653/v1/2023.emnlp-main.495}.
\newblock URL \url{https://aclanthology.org/2023.emnlp-main.495/}.

\bibitem[LangChain(2025)]{langchain2025odr}
LangChain.
\newblock Open deep research, 2025.
\newblock URL \url{https://blog.langchain.com/open-deep-research/}.

\bibitem[Lee et~al.(2025{\natexlab{a}})Lee, Kim, Lee, Moon, Hwang, Kim, Neubig,
  Welleck, and Choi]{lee2025refinebench}
Young-Jun Lee, Seungone Kim, Byung-Kwan Lee, Minkyeong Moon, Yechan Hwang,
  Jong~Myoung Kim, Graham Neubig, Sean Welleck, and Ho-Jin Choi.
\newblock Refinebench: Evaluating refinement capability in language models.
\newblock In \emph{First Workshop on Multi-Turn Interactions in Large Language
  Models}, 2025{\natexlab{a}}.
\newblock URL \url{https://openreview.net/forum?id=Ycred6ETQR}.

\bibitem[Lee et~al.(2025{\natexlab{b}})Lee, Kim, Kim, Cho, Kang, Kang, and
  Kim]{lee2025checkeval}
Yukyung Lee, JoongHoon Kim, Jaehee Kim, Hyowon Cho, Jaewook Kang, Pilsung Kang,
  and Najoung Kim.
\newblock {C}heck{E}val: A reliable {LLM}-as-a-judge framework for evaluating
  text generation using checklists.
\newblock In Christos Christodoulopoulos, Tanmoy Chakraborty, Carolyn Rose, and
  Violet Peng, editors, \emph{Proceedings of the 2025 Conference on Empirical
  Methods in Natural Language Processing}, pages 15782--15809, Suzhou, China,
  November 2025{\natexlab{b}}. Association for Computational Linguistics.
\newblock ISBN 979-8-89176-332-6.
\newblock \doi{10.18653/v1/2025.emnlp-main.796}.
\newblock URL \url{https://aclanthology.org/2025.emnlp-main.796/}.

\bibitem[Li et~al.(2025{\natexlab{a}})Li, Zeng, Cheng, Ma, and
  Jia]{li2025reportbench}
Minghao Li, Ying Zeng, Zhihao Cheng, Cong Ma, and Kai Jia.
\newblock Reportbench: Evaluating deep research agents via academic survey
  tasks, 2025{\natexlab{a}}.
\newblock URL \url{https://arxiv.org/abs/2508.15804}.

\bibitem[Li et~al.(2025{\natexlab{b}})Li, Jin, Dong, Qian, Wu, Wen, Zhu, and
  Dou]{li2025webthinker}
Xiaoxi Li, Jiajie Jin, Guanting Dong, Hongjin Qian, Yongkang Wu, Ji-Rong Wen,
  Yutao Zhu, and Zhicheng Dou.
\newblock Webthinker: Empowering large reasoning models with deep research
  capability, 2025{\natexlab{b}}.
\newblock URL \url{https://arxiv.org/abs/2504.21776}.

\bibitem[Liu et~al.(2023)Liu, Zhang, and Liang]{liu2023evaluating}
Nelson Liu, Tianyi Zhang, and Percy Liang.
\newblock Evaluating verifiability in generative search engines.
\newblock In Houda Bouamor, Juan Pino, and Kalika Bali, editors, \emph{Findings
  of the Association for Computational Linguistics: EMNLP 2023}, pages
  7001--7025, Singapore, December 2023. Association for Computational
  Linguistics.
\newblock \doi{10.18653/v1/2023.findings-emnlp.467}.
\newblock URL \url{https://aclanthology.org/2023.findings-emnlp.467/}.

\bibitem[Madaan et~al.(2023)Madaan, Tandon, Gupta, Hallinan, Gao, Wiegreffe,
  Alon, Dziri, Prabhumoye, Yang, Gupta, Majumder, Hermann, Welleck,
  Yazdanbakhsh, and Clark]{madaan2023self}
Aman Madaan, Niket Tandon, Prakhar Gupta, Skyler Hallinan, Luyu Gao, Sarah
  Wiegreffe, Uri Alon, Nouha Dziri, Shrimai Prabhumoye, Yiming Yang, Shashank
  Gupta, Bodhisattwa~Prasad Majumder, Katherine Hermann, Sean Welleck, Amir
  Yazdanbakhsh, and Peter Clark.
\newblock Self-refine: Iterative refinement with self-feedback.
\newblock In \emph{Thirty-seventh Conference on Neural Information Processing
  Systems}, 2023.
\newblock URL \url{https://openreview.net/forum?id=S37hOerQLB}.

\bibitem[Mialon et~al.(2023)Mialon, Fourrier, Swift, Wolf, LeCun, and
  Scialom]{mialon2023gaia}
Grégoire Mialon, Clémentine Fourrier, Craig Swift, Thomas Wolf, Yann LeCun,
  and Thomas Scialom.
\newblock Gaia: a benchmark for general ai assistants, 2023.
\newblock URL \url{https://arxiv.org/abs/2311.12983}.

\bibitem[Muennighoff et~al.(2025)Muennighoff, Yang, Shi, Li, Fei-Fei,
  Hajishirzi, Zettlemoyer, Liang, Candes, and Hashimoto]{muennighoff2025s1}
Niklas Muennighoff, Zitong Yang, Weijia Shi, Xiang~Lisa Li, Li~Fei-Fei,
  Hannaneh Hajishirzi, Luke Zettlemoyer, Percy Liang, Emmanuel Candes, and
  Tatsunori Hashimoto.
\newblock s1: Simple test-time scaling.
\newblock In Christos Christodoulopoulos, Tanmoy Chakraborty, Carolyn Rose, and
  Violet Peng, editors, \emph{Proceedings of the 2025 Conference on Empirical
  Methods in Natural Language Processing}, pages 20286--20332, Suzhou, China,
  November 2025. Association for Computational Linguistics.
\newblock ISBN 979-8-89176-332-6.
\newblock \doi{10.18653/v1/2025.emnlp-main.1025}.
\newblock URL \url{https://aclanthology.org/2025.emnlp-main.1025/}.

\bibitem[Nathani et~al.(2023)Nathani, Wang, Pan, and Wang]{nathani2023maf}
Deepak Nathani, David Wang, Liangming Pan, and William Wang.
\newblock {MAF}: Multi-aspect feedback for improving reasoning in large
  language models.
\newblock In Houda Bouamor, Juan Pino, and Kalika Bali, editors,
  \emph{Proceedings of the 2023 Conference on Empirical Methods in Natural
  Language Processing}, pages 6591--6616, Singapore, December 2023. Association
  for Computational Linguistics.
\newblock \doi{10.18653/v1/2023.emnlp-main.407}.
\newblock URL \url{https://aclanthology.org/2023.emnlp-main.407/}.

\bibitem[OpenAI(2025)]{openai2025deepresearch}
OpenAI.
\newblock Deep research system card, 2025.
\newblock URL \url{https://openai.com/index/deep-research-system-card/}.

\bibitem[Perplexity(2025)]{perplexity2025sonar}
Perplexity.
\newblock Introducing perplexity deep research, 2025.
\newblock URL
  \url{https://www.perplexity.ai/hub/blog/introducing-perplexity-deep-research}.

\bibitem[Phan et~al.(2025)Phan, Gatti, Han, Li, Hu, Zhang, Zhang, Shaaban,
  Ling, Shi, Choi, Agrawal, Chopra, Khoja, Kim, Ren, Hausenloy, Zhang, Mazeika,
  Yue, Wang, Hendrycks, and dataset contributors]{phan2025hle}
Long Phan, Alice Gatti, Ziwen Han, Nathaniel Li, Josephina Hu, Hugh Zhang, Chen
  Bo~Calvin Zhang, Mohamed Shaaban, John Ling, Sean Shi, Michael Choi, Anish
  Agrawal, Arnav Chopra, Adam Khoja, Ryan Kim, Richard Ren, Jason Hausenloy,
  Oliver Zhang, Mantas Mazeika, Summer Yue, Alexandr Wang, Dan Hendrycks, and
  dataset contributors.
\newblock Humanity's last exam, 2025.
\newblock URL \url{https://arxiv.org/abs/2501.14249}.

\bibitem[Qiao et~al.(2025)Qiao, Chen, Chen, Yu, Yin, Wang, Zhang, Li, Yin, Li,
  et~al.]{qiao2025webresearcher}
Zile Qiao, Guoxin Chen, Xuanzhong Chen, Donglei Yu, Wenbiao Yin, Xinyu Wang,
  Zhen Zhang, Baixuan Li, Huifeng Yin, Kuan Li, et~al.
\newblock Webresearcher: Unleashing unbounded reasoning capability in
  long-horizon agents.
\newblock \emph{arXiv preprint arXiv:2509.13309}, 2025.

\bibitem[Ruan et~al.(2025)Ruan, Nair, Cao, Liu, Munir, Pollens-Dempsey, Chiang,
  Kates, David, Chen, Yang, Yang, Gump, Bialek, Sankaran, Schlanger, and
  Wang]{ruan2025expertlongbench}
Jie Ruan, Inderjeet Nair, Shuyang Cao, Amy Liu, Sheza Munir, Micah
  Pollens-Dempsey, Tiffany Chiang, Lucy Kates, Nicholas David, Sihan Chen,
  Ruxin Yang, Yuqian Yang, Jasmine Gump, Tessa Bialek, Vivek Sankaran, Margo
  Schlanger, and Lu~Wang.
\newblock Expertlongbench: Benchmarking language models on expert-level
  long-form generation tasks with structured checklists, 2025.
\newblock URL \url{https://arxiv.org/abs/2506.01241}.

\bibitem[Shao et~al.(2025)Shao, Asai, Shen, Ivison, Kishore, Zhuo, Zhao, Park,
  Finlayson, Sontag, Murray, Min, Dasigi, Soldaini, Brahman, tau Yih, Wu,
  Zettlemoyer, Kim, Hajishirzi, and Koh]{shao2025drtulu}
Rulin Shao, Akari Asai, Shannon~Zejiang Shen, Hamish Ivison, Varsha Kishore,
  Jingming Zhuo, Xinran Zhao, Molly Park, Samuel~G. Finlayson, David Sontag,
  Tyler Murray, Sewon Min, Pradeep Dasigi, Luca Soldaini, Faeze Brahman, Wen
  tau Yih, Tongshuang Wu, Luke Zettlemoyer, Yoon Kim, Hannaneh Hajishirzi, and
  Pang~Wei Koh.
\newblock Dr tulu: Reinforcement learning with evolving rubrics for deep
  research, 2025.
\newblock URL \url{https://arxiv.org/abs/2511.19399}.

\bibitem[Sharma et~al.(2025)Sharma, Zhang, Bandi, Wang, Aich, Nghiem, Rabbani,
  Htet, Jang, Basu, Balwani, Peskoff, Ayestaran, Hendryx, Kenstler, and
  Liu]{sharma2025researchrubrics}
Manasi Sharma, Chen Bo~Calvin Zhang, Chaithanya Bandi, Clinton Wang, Ankit
  Aich, Huy Nghiem, Tahseen Rabbani, Ye~Htet, Brian Jang, Sumana Basu,
  Aishwarya Balwani, Denis Peskoff, Marcos Ayestaran, Sean~M. Hendryx, Brad
  Kenstler, and Bing Liu.
\newblock Researchrubrics: A benchmark of prompts and rubrics for evaluating
  deep research agents, 2025.
\newblock URL \url{https://arxiv.org/abs/2511.07685}.

\bibitem[Shinn et~al.(2023)Shinn, Cassano, Gopinath, Narasimhan, and
  Yao]{shinn2023reflexion}
Noah Shinn, Federico Cassano, Ashwin Gopinath, Karthik~R Narasimhan, and Shunyu
  Yao.
\newblock Reflexion: language agents with verbal reinforcement learning.
\newblock In \emph{Thirty-seventh Conference on Neural Information Processing
  Systems}, 2023.
\newblock URL \url{https://openreview.net/forum?id=vAElhFcKW6}.

\bibitem[Snell et~al.(2025)Snell, Lee, Xu, and Kumar]{snell2025scaling}
Charlie~Victor Snell, Jaehoon Lee, Kelvin Xu, and Aviral Kumar.
\newblock Scaling {LLM} test-time compute optimally can be more effective than
  scaling parameters for reasoning.
\newblock In \emph{The Thirteenth International Conference on Learning
  Representations}, 2025.
\newblock URL \url{https://openreview.net/forum?id=4FWAwZtd2n}.

\bibitem[Song et~al.(2024)Song, Kim, and Iyyer]{song2024veriscore}
Yixiao Song, Yekyung Kim, and Mohit Iyyer.
\newblock {V}eri{S}core: Evaluating the factuality of verifiable claims in
  long-form text generation.
\newblock In Yaser Al-Onaizan, Mohit Bansal, and Yun-Nung Chen, editors,
  \emph{Findings of the Association for Computational Linguistics: EMNLP 2024},
  pages 9447--9474, Miami, Florida, USA, November 2024. Association for
  Computational Linguistics.
\newblock \doi{10.18653/v1/2024.findings-emnlp.552}.
\newblock URL \url{https://aclanthology.org/2024.findings-emnlp.552/}.

\bibitem[Team et~al.(2025)Team, Li, Zhang, Zhang, Huang, Li, Chen, Yin, Wu,
  Zhou, Li, Su, Ou, Zhang, Xie, Ye, Yin, Yu, Wang, Wu, Chen, Zhao, Zhang, Tao,
  Zhang, Qiao, Wang, Yu, Fu, Shen, Yang, Lin, Zhang, Zeng, Yang, Yin, Song,
  Yan, Liao, Xia, Xiao, Min, Ding, Fang, Chen, Huang, Wang, Cai, Shen, Wang,
  Guan, Geng, Shi, Wu, Chen, Li, and Jiang]{team2025tongyi}
Tongyi~DeepResearch Team, Baixuan Li, Bo~Zhang, Dingchu Zhang, Fei Huang,
  Guangyu Li, Guoxin Chen, Huifeng Yin, Jialong Wu, Jingren Zhou, Kuan Li,
  Liangcai Su, Litu Ou, Liwen Zhang, Pengjun Xie, Rui Ye, Wenbiao Yin, Xinmiao
  Yu, Xinyu Wang, Xixi Wu, Xuanzhong Chen, Yida Zhao, Zhen Zhang, Zhengwei Tao,
  Zhongwang Zhang, Zile Qiao, Chenxi Wang, Donglei Yu, Gang Fu, Haiyang Shen,
  Jiayin Yang, Jun Lin, Junkai Zhang, Kui Zeng, Li~Yang, Hailong Yin, Maojia
  Song, Ming Yan, Minpeng Liao, Peng Xia, Qian Xiao, Rui Min, Ruixue Ding,
  Runnan Fang, Shaowei Chen, Shen Huang, Shihang Wang, Shihao Cai, Weizhou
  Shen, Xiaobin Wang, Xin Guan, Xinyu Geng, Yingcheng Shi, Yuning Wu, Zhuo
  Chen, Zijian Li, and Yong Jiang.
\newblock Tongyi deepresearch technical report, 2025.
\newblock URL \url{https://arxiv.org/abs/2510.24701}.

\bibitem[Wadhwa et~al.(2024)Wadhwa, Zhao, Li, and
  Durrett]{wadhwa2024learningrefine}
Manya Wadhwa, Xinyu Zhao, Junyi~Jessy Li, and Greg Durrett.
\newblock Learning to refine with fine-grained natural language feedback.
\newblock In Yaser Al-Onaizan, Mohit Bansal, and Yun-Nung Chen, editors,
  \emph{Findings of the Association for Computational Linguistics: EMNLP 2024},
  pages 12281--12308, Miami, Florida, USA, November 2024. Association for
  Computational Linguistics.
\newblock \doi{10.18653/v1/2024.findings-emnlp.716}.
\newblock URL \url{https://aclanthology.org/2024.findings-emnlp.716/}.

\bibitem[Wang et~al.(2025)Wang, Ming, Dulepet, Chen, Xu, Ke, Sala,
  Albarghouthi, Xiong, and Joty]{wang2025liveresearchbench}
Jiayu Wang, Yifei Ming, Riya Dulepet, Qinglin Chen, Austin Xu, Zixuan Ke,
  Frederic Sala, Aws Albarghouthi, Caiming Xiong, and Shafiq Joty.
\newblock Liveresearchbench: A live benchmark for user-centric deep research in
  the wild, 2025.
\newblock URL \url{https://arxiv.org/abs/2510.14240}.

\bibitem[Wei et~al.(2025{\natexlab{a}})Wei, Sun, Papay, McKinney, Han, Fulford,
  Chung, Passos, Fedus, and Glaese]{wei2025browsecomp}
Jason Wei, Zhiqing Sun, Spencer Papay, Scott McKinney, Jeffrey Han, Isa
  Fulford, Hyung~Won Chung, Alex~Tachard Passos, William Fedus, and Amelia
  Glaese.
\newblock Browsecomp: A simple yet challenging benchmark for browsing agents,
  2025{\natexlab{a}}.
\newblock URL \url{https://arxiv.org/abs/2504.12516}.

\bibitem[Wei et~al.(2025{\natexlab{b}})Wei, Wen, Qiao, Sun, and
  Ma]{wei2025rocketeval}
Tianjun Wei, Wei Wen, Ruizhi Qiao, Xing Sun, and Jianghong Ma.
\newblock Rocketeval: Efficient automated {LLM} evaluation via grading
  checklist.
\newblock In \emph{The Thirteenth International Conference on Learning
  Representations}, 2025{\natexlab{b}}.
\newblock URL \url{https://openreview.net/forum?id=zJjzNj6QUe}.

\bibitem[Xu et~al.(2025)Xu, Lu, Ye, Hu, and Liu]{xu2025researcherbench}
Tianze Xu, Pengrui Lu, Lyumanshan Ye, Xiangkun Hu, and Pengfei Liu.
\newblock Researcherbench: Evaluating deep ai research systems on the frontiers
  of scientific inquiry, 2025.
\newblock URL \url{https://arxiv.org/abs/2507.16280}.

\bibitem[Yang et~al.(2018)Yang, Qi, Zhang, Bengio, Cohen, Salakhutdinov, and
  Manning]{yang2018hotpotqa}
Zhilin Yang, Peng Qi, Saizheng Zhang, Yoshua Bengio, William Cohen, Ruslan
  Salakhutdinov, and Christopher~D. Manning.
\newblock {H}otpot{QA}: A dataset for diverse, explainable multi-hop question
  answering.
\newblock In Ellen Riloff, David Chiang, Julia Hockenmaier, and Jun{'}ichi
  Tsujii, editors, \emph{Proceedings of the 2018 Conference on Empirical
  Methods in Natural Language Processing}, pages 2369--2380, Brussels, Belgium,
  October-November 2018. Association for Computational Linguistics.
\newblock \doi{10.18653/v1/D18-1259}.
\newblock URL \url{https://aclanthology.org/D18-1259/}.

\bibitem[Yao et~al.(2023)Yao, Zhao, Yu, Du, Shafran, Narasimhan, and
  Cao]{yao2023react}
Shunyu Yao, Jeffrey Zhao, Dian Yu, Nan Du, Izhak Shafran, Karthik~R Narasimhan,
  and Yuan Cao.
\newblock React: Synergizing reasoning and acting in language models.
\newblock In \emph{The Eleventh International Conference on Learning
  Representations}, 2023.
\newblock URL \url{https://openreview.net/forum?id=WE_vluYUL-X}.

\bibitem[Yao et~al.(2025)Yao, Wang, Zhang, Lu, Gu, Li, Zhao, Wu, Wang, Nie,
  Teng, and Wang]{yao2025rigorousbench}
Yang Yao, Yixu Wang, Yuxuan Zhang, Yi~Lu, Tianle Gu, Lingyu Li, Dingyi Zhao,
  Keming Wu, Haozhe Wang, Ping Nie, Yan Teng, and Yingchun Wang.
\newblock {Dr. Bench}: A multidimensional evaluation for deep research agents,
  from answers to reports, 2025.
\newblock URL \url{https://arxiv.org/abs/2510.02190}.

\bibitem[Ye et~al.(2024)Ye, Sun, Arik, and Pfister]{ye2024effective}
Xi~Ye, Ruoxi Sun, Sercan Arik, and Tomas Pfister.
\newblock Effective large language model adaptation for improved grounding and
  citation generation.
\newblock In Kevin Duh, Helena Gomez, and Steven Bethard, editors,
  \emph{Proceedings of the 2024 Conference of the North American Chapter of the
  Association for Computational Linguistics: Human Language Technologies
  (Volume 1: Long Papers)}, pages 6237--6251, Mexico City, Mexico, June 2024.
  Association for Computational Linguistics.
\newblock \doi{10.18653/v1/2024.naacl-long.346}.
\newblock URL \url{https://aclanthology.org/2024.naacl-long.346/}.

\bibitem[Zelikman et~al.(2024)Zelikman, Lorch, Mackey, and
  Kalai]{zelikman2024self}
Eric Zelikman, Eliana Lorch, Lester Mackey, and Adam~Tauman Kalai.
\newblock Self-taught optimizer ({STOP}): Recursively self-improving code
  generation.
\newblock In \emph{First Conference on Language Modeling}, 2024.
\newblock URL \url{https://openreview.net/forum?id=46Zgqo4QIU}.

\bibitem[Zhou et~al.(2023)Zhou, Lu, Mishra, Brahma, Basu, Luan, Zhou, and
  Hou]{zhou2023ifeval}
Jeffrey Zhou, Tianjian Lu, Swaroop Mishra, Siddhartha Brahma, Sujoy Basu,
  Yi~Luan, Denny Zhou, and Le~Hou.
\newblock Instruction-following evaluation for large language models, 2023.
\newblock URL \url{https://arxiv.org/abs/2311.07911}.

\end{thebibliography}
